\documentclass[runningheads]{llncs}
\usepackage{graphicx}
\usepackage{amsmath,amssymb} 
\usepackage{color}

\usepackage{color, colortbl}
\usepackage{hyperref}
\usepackage{wrapfig}
\usepackage{subcaption}
\usepackage{makecell}
\usepackage{multirow}
\usepackage{ dsfont }
\usepackage{pifont}
\newcommand{\cmark}{\ding{51}}%
\newcommand{\xmark}{\ding{55}}%
\definecolor{darkBlue}{rgb}{0.5,0.6,1}
\definecolor{LightCyan}{rgb}{0.5,0.8,1}
\definecolor{yellow}{rgb}{1,1,0.6}

\definecolor{green}{rgb}{0,0.4,0.2}
\definecolor{red}{rgb}{0.8,0,0.1}

\newcommand{\R}{\mathbb{R}}

\newcommand{\beginsupplement}{%
        \setcounter{table}{0}
        \setcounter{figure}{0}
        \setcounter{section}{0}
     }

\begin{document}
\pagestyle{headings}
\mainmatter

\def\ACCV20SubNumber{297}  

\title{To Filter Prune, or to Layer Prune, That Is The Question} 
\titlerunning{To Filter Prune, or to Layer Prune, That Is The Question}
%
\author{Sara Elkerdawy\inst{1}\orcidID{0000-0002-9607-3225} \and
Mostafa Elhoushi\inst{2} \and
Abhineet Singh\inst{1} \and
Hong Zhang\inst{1}\and
Nilanjan Ray\inst{1}}
\authorrunning{S. Elkerdawy et al.}
%
\institute{Department of Computing Science, University of Alberta, Canada\\ \email{\{elkerdaw, asingh1, hzhang, nray1\}@ualberta.ca}
\and
Toronto Heterogeneous Compilers Lab, Huawei, Canada}

\maketitle

\setcounter{footnote}{0} 


\begin{abstract}
Recent advances in pruning of neural networks have made it possible to remove a large number of filters or weights without any perceptible drop in accuracy. The number of parameters and that of FLOPs are usually the reported metrics to measure the quality of the pruned models. However, the gain in speed for these pruned models is often overlooked in the literature due to the complex nature of latency measurements. In this paper, we show the limitation of filter pruning methods in terms of latency reduction and propose LayerPrune framework. LayerPrune presents a set of layer pruning methods based on different criteria that achieve higher latency reduction than filter pruning methods on similar accuracy. The advantage of layer pruning over filter pruning in terms of latency reduction is a result of the fact that the former is not constrained by the original model's depth and thus allows for a larger range of latency reduction. For each filter pruning method we examined, we use the same filter importance criterion to calculate a per-layer importance score in one-shot. We then prune the least important layers and fine-tune the shallower model which obtains comparable or better accuracy than its filter-based pruning counterpart. This one-shot process allows to remove layers from single path networks like VGG before fine-tuning, unlike in iterative filter pruning, a minimum number of filters per layer is required to allow for data flow which constraint the search space. To the best of our knowledge, we are the first to examine the effect of pruning methods on latency metric instead of FLOPs for multiple networks, datasets and hardware targets. LayerPrune also outperforms handcrafted architectures such as Shufflenet, MobileNet, MNASNet and ResNet18 by 7.3\%, 4.6\%, 2.8\% and 0.5\% respectively on similar latency budget on ImageNet dataset. \footnote{Code is available at \url{https://github.com/selkerdawy/filter-vs-layer-pruning}}




\end{abstract}


\begin{figure}[t]
\begin{center}

\includegraphics[width=0.7\linewidth]{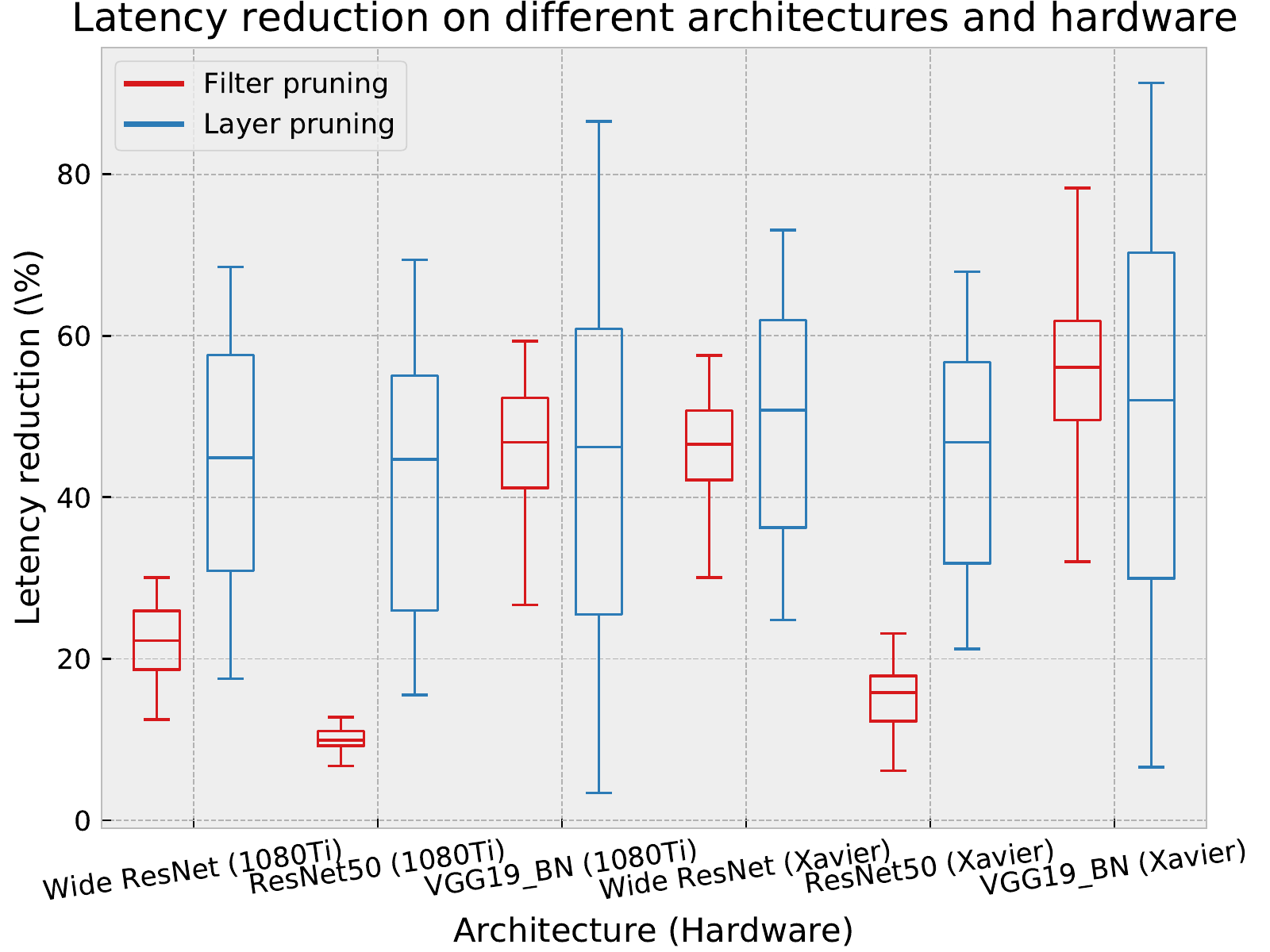}
\end{center}
\caption{Example of 100 randomly pruned models per boxplot generated from different architectures. The plot shows layer pruned models have a wider range of attainable latency reduction consistently across architectures and different hardware platforms (1080Ti and Xavier). Latency is estimated using 224x224 input image and batch size=1.}

\label{fig:intro}
\end{figure}

\section{Introduction}\label{intro}

Convolutional Neural Networks (CNN) have become the state-of-the-art in various computer vision tasks, e.g., image classification \cite{krizhevsky2012imagenet}, object detection \cite{redmon2018yolov3}, depth estimation \cite{elkerdawy2019lightweight}. These CNN models are designed with deeper \cite{he2016deep} and wider \cite{wu2019wider} convolutional layers with a large number of parameters and convolutional operations. These architectures hinder deployment on low-power devices, e.g, phones, robots, wearable devices as well as real-time critical applications, such as autonomous driving. As a result, computationally efficient models are becoming increasingly important and multiple paradigms have been proposed to minimize the complexity of CNNs.

A straight forward direction is to manually design networks with a small footprint from the start such as \cite{ma2018shufflenet,howard2017mobilenets,wang2018pelee,huang2017densely,huang2017multi}. This direction does not only require expert knowledge and multiple trials (e.g up to 1000 neural architectures explored manually \cite{darkarts}), but also does not benefit from available, pre-trained large models. Quantization \cite{wang2019haq,hubara2017quantized} and distillation \cite{yang2019snapshot,jin2019knowledge} are two other techniques, which utilize the pre-trained models to obtain smaller architectures. Quantization reduces bit-width of parameters and thus decreases memory footprint, but requires specialized hardware instructions to achieve latency reduction. While distillation trains a pre-defined smaller model (student) with guidance from a larger pre-trained model (teacher) \cite{yang2019snapshot}. Finally, model pruning aims to automatically remove the least important filters (or weights) to reduce the number of parameters or FLOPs (i.e indirect measures). However, prior work \cite{yang2017designing,yang2018netadapt,bianco2018benchmark} showed that neither number of pruned parameters nor FLOPs reduction directly correlate with latency (i.e a direct measure) consumption. Latency reduction, in that case, depends on various aspects, such as the number of filters per layer (signature) and the deployment device. Most GPU programming tools require careful compute kernels\footnote{A compute kernel refers to a function such as convolution operation that runs on a high throughput accelerator such as GPU} tuning for different matrices shapes (e.g., convolution weights) \cite{van2019kernel,nugteren2015cltune}. These aspects introduce non-linearity in modeling latency with respect to the number of filters per layer. Recognizing the limitations in terms of latency or energy by simply pruning away filters, recent works \cite{yang2018netadapt,ecc,yang2017designing} proposed optimizing directly over these direct measures. These methods require per hardware and architecture latency measurements collection to create lookup-tables or latency prediction models which can be time-intensive. In addition, these filter pruned methods are bounded by the model's depth and can only reach a limited goal for latency consumption.

In this work, we show the limitations of filter pruning methods in terms of latency reduction. Fig. \ref{fig:intro} shows the range of attainable latency reduction on randomly generated models. Each box bar summarizes the latency reduction of 100 random models with filter and layer pruning on different network architectures and hardware platforms. For each filter pruned model $i$, a pruning ratio $p_{i,j}$ per layer $j$ such that $0 \leq p(i,j)\leq 0.9$ is generated thus models differ in signature/width. For each layer pruned model, $M$ layers out of total $L$ layers (dependent on the network) are randomly selected for retention such that $1 \leq M \leq L$ thus models differ in depth. As to be expected, layer pruning has a higher upper bound in latency reduction compared to filter pruning especially on modern complex architectures with residual blocks. However, we want to highlight quantitatively in the plot the discrepancy of attainable latency reduction using both methods. Filter pruning is not only constrained by the depth of the model but also by the connection dependency in the architecture. An example of such connection dependency is the element-wise sum operation in the residual block between identity connection and residual connection. Filter pruning methods commonly prune in-between convolution layers in a residual to respect the number of channels and spatial dimensions. BAR \cite{bar} proposed an atypical residual block that allows mixed-connectivity between blocks to tackle the issue. However, this requires special implementations to leverage the speedup gain. Another limitation in filter pruning is the iterative process and thus is constrained to keep a minimum number of filters per layer during optimization to allow for data passing. LayerPrune performs a one-shot pruning before fine-tuning and thus it allows for layer removal even from single path networks.

\begin{figure}[t]
\begin{center}

\includegraphics[width=0.9\linewidth]{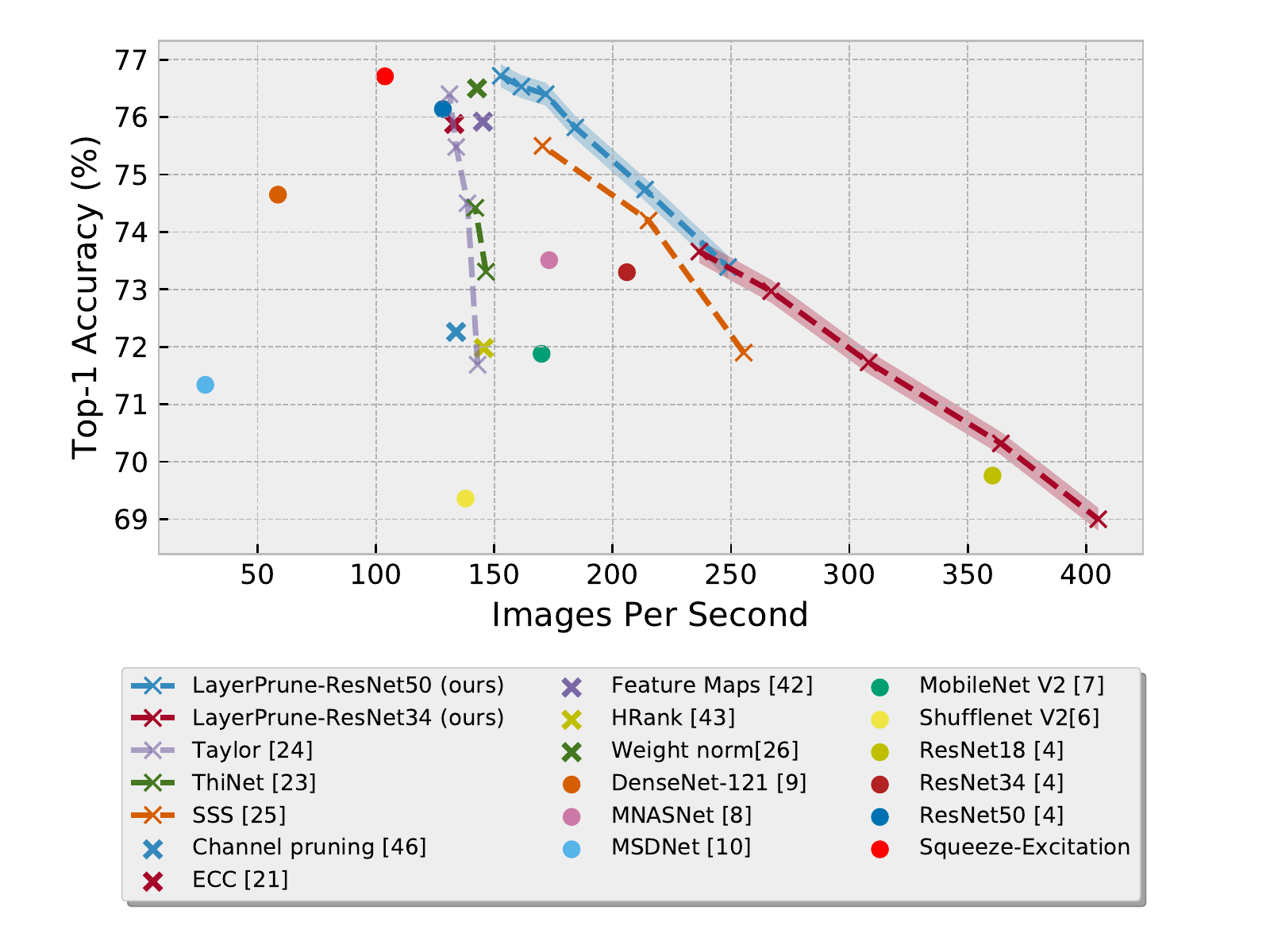}
\end{center}
\caption{Evaluation on ImageNet between our LayerPrune framework, handcrafted architectures (dots) and pruning methods on ResNet50 (crosses). Inference time is measured on 1080Ti GPU.}

\label{fig:acc}
\end{figure}

Motivated by these points, what remains to ask is how well do layer pruned models perform in terms of accuracy compared to filter pruned methods. Fig. \ref{fig:acc} shows accuracy and images per second between our LayerPrune, several state-of-the-art pruning methods, and handcrafted architectures. In general, pruning methods tend to find better quality models than handcrafted architectures. It is worth noting that filter pruning methods such as ThiNet \cite{luo2017thinet} and Taylor \cite{taylor} show small speedup gain as more filters are pruned compared to LayerPrune. That shows the limitation of filter pruning methods on latency reduction.


\section{Related Work}

We divide existing pruning methods into four categories: weight pruning, hardware-agnostic filter pruning, hardware-aware filter pruning and layer pruning.\\

\textbf{Weight pruning}. An early major category in pruning is individual weight pruning (unstructured pruning). Weight pruning methods leverage the fact that some weights have minimal effect on the task accuracy and thus can be zeroed-out. In \cite{han2015learning}, weights with small magnitude are removed and in \cite{han2015compressing}, quantization is further applied to achieve more model compression. Another data-free pruning is \cite{srinivas2015data} where neurons are removed iteratively from fully connected layers. $L_0$-regularization based method \cite{louizos2017learning} is proposed to encourage network sparsity in training. Finally, in lottery ticket hypothesis \cite{frankle2018lottery}, the authors propose a method of finding winning tickets which are subnetworks from random initialization that achieve higher accuracy than the dense model. The limitation of the unstructured weight pruning is that dedicated hardware and libraries \cite{sharify2019laconic} are needed to achieve speedup from the compression. Given our focus on latency and to keep the evaluation setup simple, we do not consider these methods in our evaluation.\\

\textbf{Hardware-agnostic filter pruning}. Methods in this category (also known as structured pruning) aim to reduce the footprint of a model by pruning filters without any knowledge of the inference resource consumption. Examples of these are \cite{taylor,slimming,luo2017thinet,wen2016learning,Li2016PruningFF}, which focus on removing the least important filters and obtaining a slimmer model. Earlier filter-pruning methods \cite{luo2017thinet,Li2016PruningFF} required layer-wise sensitivity analysis to generate the signature (i.e number of filters per layer) as a prior and remove filters based on a filter criterion. The sensitivity analysis is computationally expensive to conduct and becomes even less feasible for deeper models. Recent methods \cite{taylor,slimming,wen2016learning} learn a global importance measure removing the need for sensitivity analysis. Molchanov et al. \cite{taylor} propose a Taylor approximation on the network's weights where the filter's gradients and norm are used to approximate its global importance score. Liu et al. \cite{slimming} and Wen et al. \cite{wen2016learning} propose sparsity loss for training along with the classification's cross-entropy loss. Filters whose criterion are less than a threshold are removed and the pruned model is finally fine-tuned. Zhao et al. \cite{zhao2019variational} introduce channel saliency that is parameterized as Gaussian distribution and optimized in the training process. After training, channels with small mean and variance are pruned. In general, methods with sparsity loss lack a simple approach to respect a resource consumption target and require hyperparameter tuning to balance different losses. 

\textbf{Hardware-aware filter pruning}. To respect a resource consumption budget, recent works \cite{chin2018layer,yang2018netadapt,ecc,he2018amc} have been proposed to take into consideration a resource target within the optimization process. NetAdapt \cite{yang2018netadapt} prunes a model to meet a target budget using a heuristic greedy search. A lookup table is built for latency prediction and then multiple candidates are generated at each pruning iteration by pruning a $ratio$ of filters from each layer independently. The candidate with the highest accuracy is then selected and the process continues to the next pruning iteration with a progressively increasing $ratio$. On the other hand, AMC \cite{he2018amc} and ECC \cite{ecc} propose an end-to-end constrained pruning. AMC utilizes reinforcement learning to select a model's signature by trial and error. ECC simplifies the latency reduction model as a bilinear per-layer model. The training utilizes the alternating direction method of multipliers (ADMM) algorithm to perform constrained optimization by alternating between network weight optimization and dual variables that control the layer-wise pruning ratio. Although these methods incorporate resource consumption as a constraint in the training process, the range of attainable budgets is limited by the depth of the model. Besides, generating data measurements to model resource consumption per hardware and architecture can be expensive especially on low-end hardware platforms.

\textbf{Layer pruning}. Unlike filter pruning, little attention is paid to shallow CNNs in the pruning literature. In SSS \cite{huang2018data}, the authors propose to train a scaling factor for structure selection such as neurons, blocks, and groups. However, shallower models are only possible with architectures with residual connections to allow data flow in the optimization process. Closest to our work for a general (unconstrained by architecture type) layer pruning approach is the work done by Chen et al. \cite{chen2018shallowing}. In their method, linear classifiers probes are utilized and trained independently per layer for layer-ranking. After the layer-ranking learning stage, they prune the least important layers and fine-tune the shallower model. Although \cite{chen2018shallowing} requires rank training, it is without any gain in classification accuracy compared to our one-shot LayerPrune layer ranking as will be shown in the experiments section.


\begin{figure}[t]
\begin{center}

\includegraphics[width=0.9\linewidth]{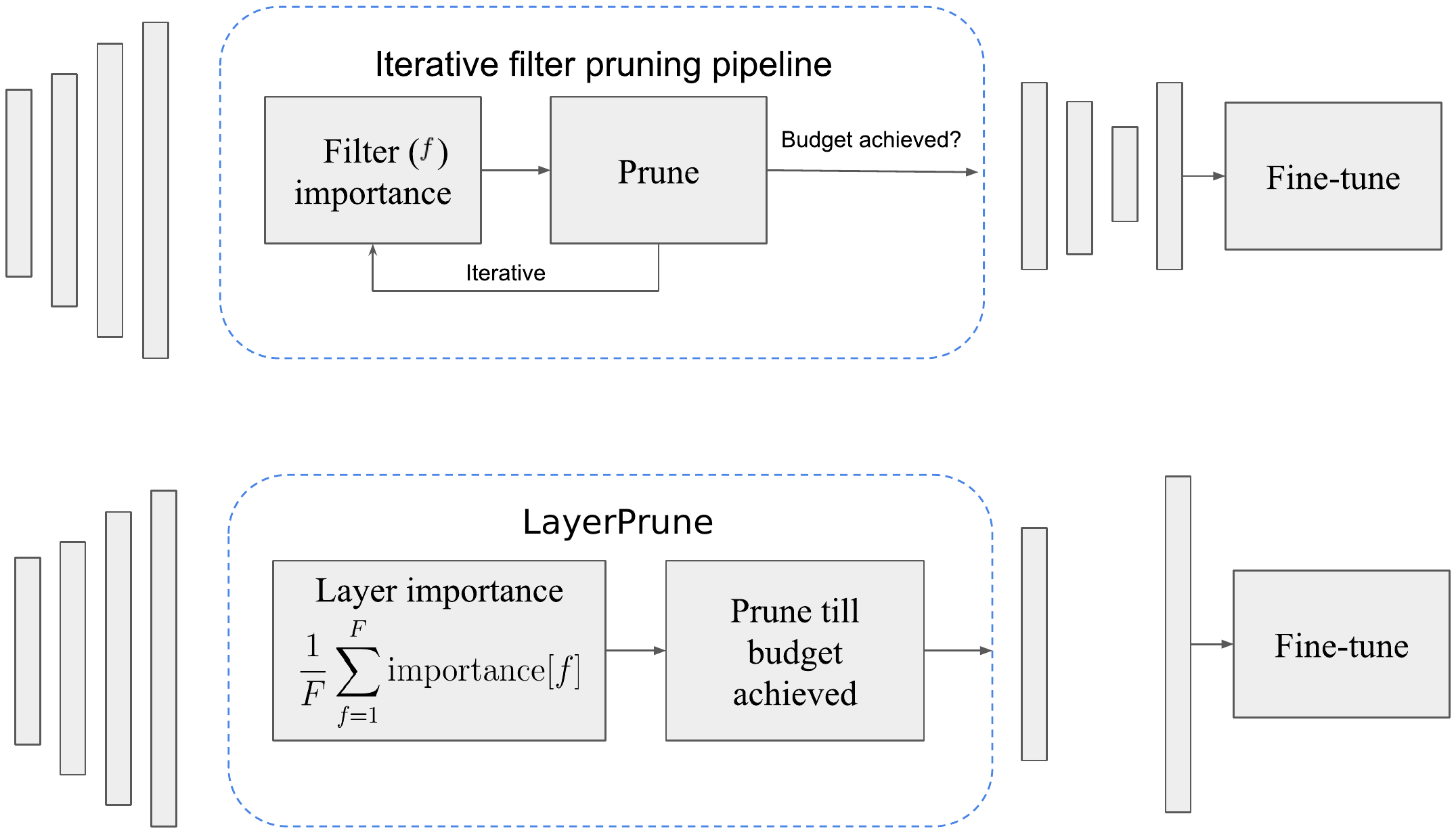}
\end{center}
\caption{Main pipeline illustrates the difference between typical iterative filter pruning and proposed LayerPrune framework. Filter pruning (top) produces thinner architecture in an iterative process while LayerPrune (bottom) prunes whole layers in one-shot. In LayerPrune, layer's importance is calculated as the average importance of each filter $f$ in all filters $F$ at that layer.}

\label{fig:pipeline}
\end{figure}

\section{Methodology}
In this section, we describe in detail LayerPrune for layer pruning using existing filter criteria along with a novel layer-wise accuracy approximation. A typical filter pruning method follows a three-stage pipeline as illustrated in Figure \ref{fig:pipeline}. Filter importance is iteratively re-evaluated after each pruning step based on a pruning meta-parameter such as pruning N filters or pruning those $\le$ threshold. In LayerPrune, we remove the need for the iterative pruning step and show that using the same filter criterion, we can remove layers in a one-shot to respect a budget. This simplifies the pruning step to a hyper-parameter free process and is computationally efficient. Layer importance is calculated as the average of filter importance in this layer. \\

\subsection{Statistics-based criterion} \label{filtercriteria}
Although existing filter pruning methods are different in algorithms and optimization used, they focus more on finding the optimal per-layer number of filters and share common filter criteria. We divide the methods based on the filter criterion used and propose their layer importance counterpart used in LayerPrune. \\

\textbf{Preliminary notion.} Consider a network with $L$ layers, each layer $l$ has weight matrix $W^{(l)} \in \mathbb{R}^{N_l\times F_l \times K_l \times K_l}$ with $N_l$ input channels, $F_l$ number of filters and $K_l$ is the size of the filters at this channel. Evaluated criteria and methods are: \\

\textbf{Weight statistics.} \cite{han2015learning,Li2016PruningFF,ecc} differ in the optimization algorithm but share weight statistics as a filter ranking. Layer pruning for this criteria is calculated as:
\begin{equation} 
\begin{gathered}
    \mbox{weights-layer-importance}[l]= \frac{1}{F_l}\sum_{i=1}^{F_l} \left\| W^{(l)}[:,i,:,:]\right \|_2
\end{gathered} 
\end{equation}

\textbf{Taylor weights.} Taylor method \cite{taylor} is slightly different from previous criterion in that the gradients are included in the ranking as well. Filter $f$ ranking is based on $\sum_{s}(g_sw_s)^2$ where $s$ iterates over all individual weights in $f$, $g$ is the gradient, $w$ is the weight value. Similarly, layer ranking can be expressed as:
\begin{equation} 
\begin{gathered}
    \mbox{taylor-layer-importance}[l]= \frac{1}{F_l}\sum_{i=1}^{F_l} \left\| G^{(l)}[:,i,:,:] \odot W^{(l)}[:,i,:,:]\right\|_2
\end{gathered} 
\end{equation}

where $\odot$ is element-wise product and $G^{(l)} \in \mathbb{R}^{N_l\times F_l \times K_l \times K_l}$ is the gradient of loss with respect to weights $W^{(l)}$.

\textbf{Feature map based heuristics.} \cite{luo2017thinet,fmtaylor,lin2020hrank} rank filters based on statistics from output of layer. In \cite{luo2017thinet}, ranking is based on the effect on the next layer while \cite{fmtaylor}, similar to Taylor weights, utilizes gradients and norm but on feature maps.

\textbf{Channel saliency.} In this criterion, a scalar is multiplied by the feature maps and optimized within a typical training cycle with task loss and sparsity regularization loss to encourage sparsity. Slimming \cite{slimming} utilizes Batch Normalization scale $\gamma$ as the channel saliency. Similarly, we use Batch Normalization scale parameter to calculate layer importance for this criteria, specifically:

\begin{equation} 
\begin{gathered}
    \mbox{BN-layer-importance}[l]= \frac{1}{F_l}\sum_{i=1}^{F_l} (\gamma_i^{(l)})^2
\end{gathered} 
\end{equation}

\textbf{Ensemble.} We also consider diverse ensemble of layer ranks where the ensemble rank of each layer is the sum of its rank per method, more specifically:

\begin{equation}
\begin{gathered}
   \mbox{ensemble-rank}[l] = \sum_{m \in \{1...M\}} (\mbox{LayerRank}(m,l))
\end{gathered} 
\end{equation}
where $l$ is the layer's index, $M$ is the number of all criteria and LayerRank indicates the order of layer $l$ in the sorted list for criterion $m$.

\subsection{Efficiency-based criterion}\label{layerprune} 

In addition to existing filter criteria, we present a novel layer importance by layer-wise accuracy approximation. Motivated by the few-shot learning literature \cite{qi2018low,imprintseg}, we use imprinting to approximate the classification accuracy up to each layer. Imprinting is used to approximate a classifier's weight matrix when only a few training samples are available. Although we have adequate training samples, we are inspired by the efficiency of imprinting to approximate the accuracy in one pass without the need for training. We create a classifier proxy for each prunable candidate (e.g convolution layer or residual blocks), and then the training data is used to imprint the classifier weight matrix for each proxy. Since each layer has a different output feature shape, we apply adaptive average pooling to simplify our method and unify the embedding length so that each layer produces roughly an output of the same size. Specifically, the pooling is done as follows:\\
\begin{equation} \label{eq:imprint}
\begin{gathered}
   d_i = \mbox{round}(\sqrt{\frac{N}{n_i}})\\
   E_i = \mbox{AdaptiveAvgPool}(O_i,d_i),
\end{gathered} 
\end{equation}
where $N$ is the embedding length, $n_i$ is layer $i$'s number of filters, $O_i$ is layer $i$'s output feature map, and AdaptiveAvgPool \cite{spp} reduces $O_i$ to embedding $ E_i \in \R^{d_i\times d_i \times n_i}$. Finally, embeddings per layer are flattened to be used in imprinting. Imprinting calculates the proxy classifier's weights matrix $P_i$ as follows:\\
\begin{equation} 
\begin{gathered}
    P_i[:,c] = \frac{1}{N_c} \sum_{j=1}^{D} \mathds{I}_{[c_j==c]} E_j
\end{gathered} 
\end{equation}
where $c$ is the class id, $c_j$ is sample's $j$ class id, $N_c$ is the number of samples in class $c$, $D$ is the total number of samples, and $\mathds{I}_{[.]}$ denotes the indicator function.

The accuracy at each proxy is then calculated using the imprinted weight matrices. The prediction for each sample $j$ is calculated for each layer $i$ as:
\begin{equation} 
\begin{gathered}
    \hat{y}_j = \operatorname*{argmax}_{c \in \{1,...,C\}} P_i[:,c]^TE_j,
\end{gathered} 
\end{equation}
where $E_j$ is calculated as shown in Eq.(\ref{eq:imprint}). This is equivalent to finding the nearest class from the imprinted weights in the embedding space. Ranking of each layer is then calculated as the gain in accuracy from previous pruning candidate.

\section{Evaluation Results}
In this section we present our experimental results comparing state-of-the-art pruning methods and LayerPrune in terms of accuracy and latency reduction on two different hardware platforms. We show latency on high-end GPU 1080Ti and on NVIDIA Jetson Xavier embedded device, which is used in mobile vision systems and contains 512-core Volta GPU. We evaluate the methods on CIFAR10/100 \cite{cifar} and ImageNet \cite{krizhevsky2012imagenet} datasets.

\subsection{Implementation details}
\textbf{Latency calculation.} Latency model is averaged over 1000 forward pass after 10 warm up forward passes for lazy GPU initialization. Latency is calculated using batch size 1, unless otherwise stated, due to its practical importance in real-time application as in robotics where we process an online stream of frames. All pruned architectures are implemented and measured using PyTorch \cite{baydin2017automatic}. For a fair comparison, we compare latency reduction on similar accuracy retention from baseline and reported by original papers or compare accuracy on similar latency reduction with methods supporting layer or block pruning.\\
\textbf{Handling filter shapes after layer removal.} If the pruned layer $l$ with weight $W^{(l)} \in \mathbb{R}^{N_l\times F_l \times K_l \times K_l}$ has $N_{l} \neq F_l$, we replace layer $(l+1)$'s weight matrix from $W^{(l+1)} \in \mathbb{R}^{F_l\times F_{l+1} \times K_{l+1} \times K_{l+1}}$ to $W^{(l+1)} \in \mathbb{R}^{N_l\times F_{l+1} \times K_{l+1} \times K_{l+1}}$ with random initialization. All other layers are initialized from the pre-trained dense model.

\subsection{Results on CIFAR}
We evaluate CIFAR-10 and CIFAR-100 on ResNet56 \cite{he2016deep} and VGG19-BN \cite{VGG}.

\subsubsection{Random filters vs. Random layers} Initial hypothesis verification is to generate random filter and layer pruned models, then train them to compare their accuracy and latency reduction. Random models generation follows the same setup as explained in Section (\ref{intro}). Each model is trained with SGD optimization for 164 epochs with learning rate 0.1 that decays by 0.1 at epochs 81, 121, and 151. Figure \ref{fig:random} shows the latency-accuracy plot for both random pruning methods. Layer pruned models outperform filter pruned ones in accuracy by 7.09\% on average and can achieve up to 60\% latency reduction. Also, within the same latency budget, filter pruning shows higher variance in accuracy than layer pruning. This suggests that latency constrained optimization with filter pruning is complex and requires careful per layer pruning ratio selection. On the other hand, layer pruning has small accuracy variation, in general within a budget.

\subsubsection{VGG19-BN} 
Results on CIFAR-100 are presented in Table \ref{tab:cifar100vgg}. The table is divided based on the previously mentioned filter criterion categorization in Section \ref{filtercriteria}. First, we compare with Chen et el. \cite{chen2018shallowing} on a similar latency reduction as both \cite{chen2018shallowing} and LayerPrune perform layer pruning. Although \cite{chen2018shallowing} requires training for layer ranking, LayerPrune outperforms it by 1.11\%. We achieve up to 56\% latency reduction with 1.52\% accuracy increase from baseline. As VGG19-BN is over-parametrized for CIFAR-100, removing layers act as a regularization and can find models with better accuracy than the baseline. Unlike with filter pruning methods, they are bounded by small accuracy variations around the baseline. It is worth mentioning that latency reduction of removing the same number of filters using different filter criteria varies from -0.06\% to 40.0\%. While layer pruned models, with the same number of pruned layers, regardless of the criterion range from 34.3\% to 41\%. That suggests that latency reduction using filter pruning is sensitive to environment setup and requires complex optimization to respect a latency budget.

\begin{figure}[t]
\centering
\begin{minipage}{.58\textwidth}
\centering
\includegraphics[width=1\linewidth]{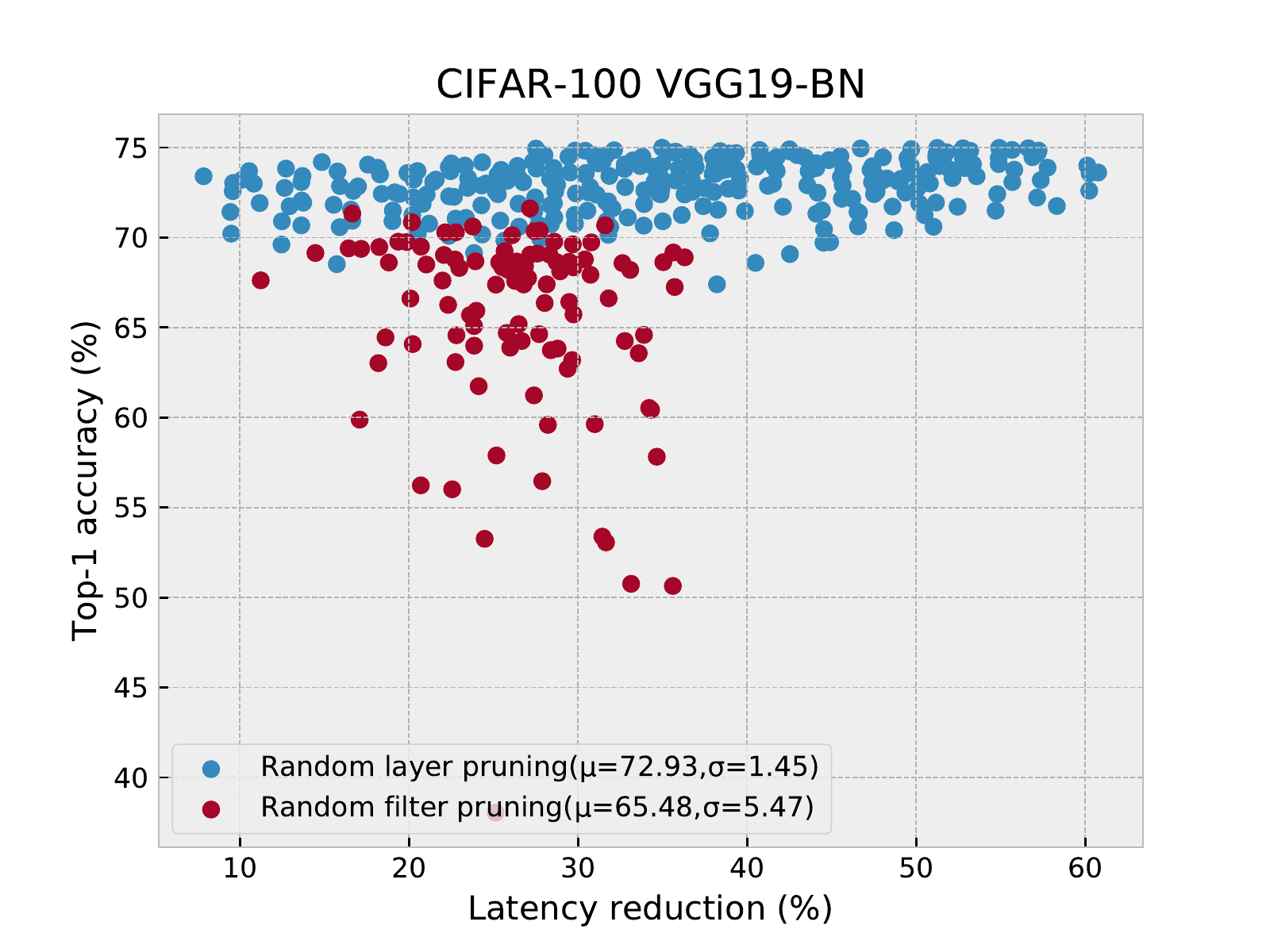}
\end{minipage}\qquad
\begin{minipage}{.36\textwidth}
\centering
\includegraphics[width=1\linewidth]{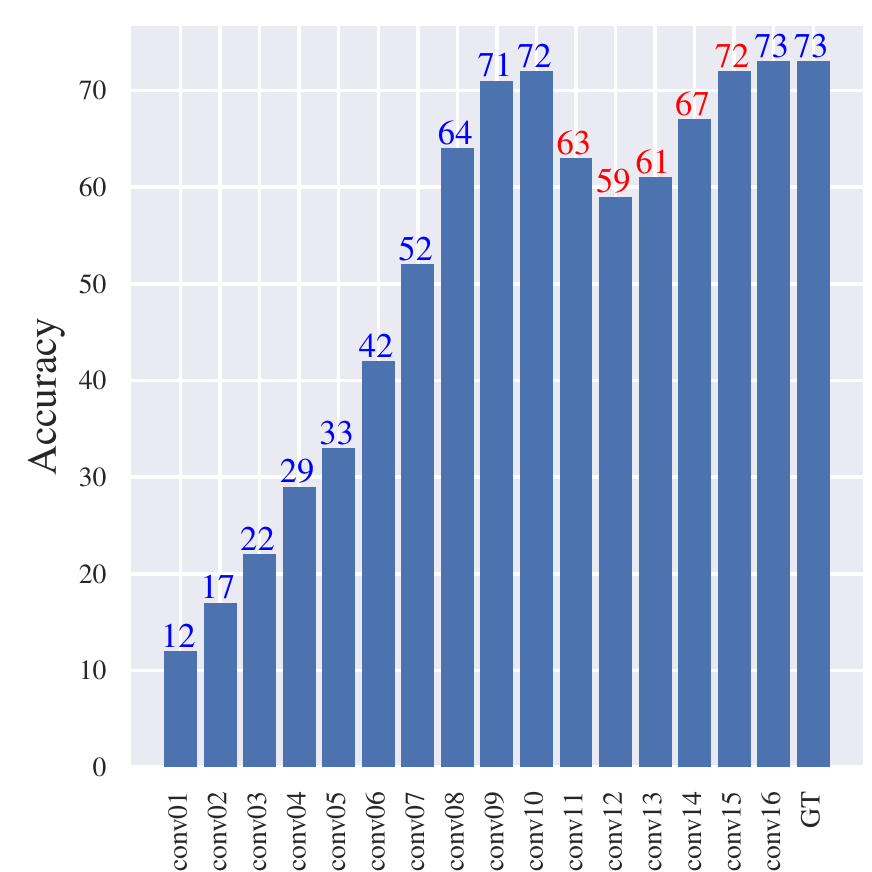}
\end{minipage}

\bigskip

\begin{minipage}[t]{.58\textwidth}
\centering

\caption{Random filter pruned and layer pruned models generated from VGG19-BN (Top-1=73.11\%). Accuracy mean and standard deviation is shown in parentheses.}
\label{fig:random}
\end{minipage}\qquad
\begin{minipage}[t]{.36\textwidth}
\centering
\caption{Layer-wise accuracy using imprinting on CIFAR-100. Red indicates drop in accuracy.}
\label{fig:vggimprint}
\end{minipage}
\end{figure}

To further explain the accuracy increase by LayerPrune, Fig. \ref{fig:vggimprint} shows layer-wise accuracy approximation on baseline VGG19-BN using the imprinting method explained in Section (\ref{layerprune}). Each bar represents the approximated classification accuracy up to this layer (rounded for visualization). We see a drop in accuracy followed by an increasing trend from conv10 to conv15. This is likely because the number of features is the same from conv10 to conv12. We start to observe an accuracy increase only at conv13 that follows a max-pooling layer and has twice as many features. That highlights the importance of downsampling and doubling the number of features at this point in the model. So layer pruning does not only improve inference speed but can also discover a better regularized shallow model especially on a small dataset. It is also worth mentioning that both the proxy classifier from the last layer, conv16, and the actual model classifier, GT, have the same accuracy, showing how the proxy classifier is a plausible approximation of the converged classifier. 

\begin{table}[t]
\centering
\setlength{\tabcolsep}{6pt}
\resizebox{1.\columnwidth}{!}{
\begin{tabular}{|@{\hspace{0.2em}}l|@{}c|c||c|c|c|c|}
\hline
\multicolumn{7}{|c|}{\textbf{VGG19 (73.11\%)}} \\
\hline
\multicolumn{1}{|@{\hspace{0.2em}}l|}{\multirow{2}{*}{Method}} & \multicolumn{1}{c|}{\multirow{2}{*}{Shallower?}} & \multicolumn{1}{c||}{\multirow{2}{*}{\begin{tabular}[c]{@{}c@{}}Top1 \\ Acc. (\%)\end{tabular}}} & \multicolumn{2}{c|}{ 1080Ti LR (\%)} & \multicolumn{2}{c|}{ Xavier LR (\%)} \\ \cline{4-7} 
\multicolumn{1}{|l|}{} & \multicolumn{1}{c|}{} & \multicolumn{1}{c||}{} & \multicolumn{1}{c|}{bs=8} & \multicolumn{1}{c|}{bs=64} & \multicolumn{1}{c|}{bs=8} & \multicolumn{1}{c|}{bs=64} \\ \hline
Chen et al. \cite{chen2018shallowing} & \cmark & 73.25 & 56.01 & 52.86 & 58.06 & 49.86 \\
LayerPrune$_8$-Imprint & \cmark & \textbf{74.36} & 56.10 & 53.67 & 57.79 & 49.10 \\
\hline
Weight norm \cite{han2015learning} & \xmark & 73.01 & -2.044 & -0.873 & -4.256 & -0.06 \\
ECC \cite{ecc} & \xmark & 72.71 & 16.37 & 36.70 & 29.17 & \textbf{36.69} \\
LayerPrune$_2$ & \cmark &73.60 & 17.32 & 14.57 & 19.512 & 10.97 \\
LayerPrune$_5$ & \cmark & \textbf{74.80} & \textbf{39.84} & \textbf{37.85} & \textbf{41.86} & 34.38 \\
\hline
Slimming \cite{slimming} & \xmark &72.32 & 16.84 & \textbf{40.08} & 40.55 & \textbf{39.53} \\
LayerPrune$_2$ & \cmark & 73.60 & 17.34 & 13.86 & 18.85 & 10.90 \\
LayerPrune$_5$ & \cmark & \textbf{74.80} & \textbf{39.56} & 37.30 & \textbf{41.40} & 34.35 \\
\hline
Taylor \cite{taylor} & \xmark &72.61 & 15.87 & 19.77 & -4.89 & 17.45 \\
LayerPrune$_2$ & \cmark &73.60 & 17.12 & 13.54 & 18.81 & 10.89 \\
LayerPrune$_5$ & \cmark &\textbf{74.80} & \textbf{39.36} & \textbf{37.12} &\textbf{ 41.34}& \textbf{34.44} \\
\hline
\end{tabular}}
  \caption{\textbf{Comparison of different pruning methods on VGG19-BN CIFAR-100.} The accuracy for baseline model is shown in parentheses. LR, bs stands for latency reduction and batch size respectively. $x$ in LayerPrune$_x$ indicates number of layers removed. -ve LR indicates increase in latency. Shallower indicates whether a method prunes layers. Best is shown in \textbf{bold}.}
  \label{tab:cifar100vgg}
\end{table}

\subsubsection{ResNet56} 
We also compare on the more complex architecture ResNet56 on CIFAR-10 and CIFAR-100 in Table \ref{tab:cifarresnet50}. On a similar latency reduction, LayerPrune outperforms \cite{chen2018shallowing} by 0.54\% and 1.23\% on CIFAR-10 and CIFAR-100 respectively. On the other hand, within each filter criterion, LayerPrune outperforms filter pruning and is on par with the baseline in accuracy. In addition, filter pruning can result in latency increase (i.e negative LR) with specific hardware targets and batch sizes \cite{sze2017efficient} as shown with batch size 8. However, LayerPrune consistently shows latency reduction under different environmental setups. We also compare with larger batch size to further encourage filter pruned models to better utilize the resources. Still, we found LayerPrune achieves overall better latency reduction with a large batch size. Latency reduction variance, LR var, between different batch sizes within the same hardware platform is shown as well. Consistent with previous results on VGG, LayerPrune is less sensitive to changes in criterion, batch size, and hardware than filter pruning. We also show results up to 2.5x latency reduction with less than 2\% accuracy drop.

\subsection{Results on ImageNet}
We evaluate the methods on ImageNet dataset for classification. For all experiments in this section, PyTorch pre-trained models are used as a starting point for network pruning. We follow the same setup as in \cite{taylor} where we prune 100 filters each 30 mini-batches for 10 pruning iterations. The pruned model is then fine-tuned with learning rate 1e$^{-3}$ using SGD optimizer and 256 batch size. Results on ResNet50 are presented in Table \ref{tab:imgnetresnet50}. In general, LayerPrune models improve accuracy over the baseline and their counterpart filter pruning methods. Although feature maps criterion \cite{fmtaylor} achieves better accuracy by 0.92\% over LayerPrune$_1$, LayerPrune has higher latency reduction that exceeds by 5.7\%. 
\clearpage

\begin{table}[!htbp]

\centering
\setlength{\tabcolsep}{6pt}
\resizebox{1.\columnwidth}{!}{
\begin{tabular}{|@{\hspace{0.2em}}l|c|c||c|c|c|c|}
\hline
\multicolumn{1}{|@{\hspace{0.2em}}l|}{\multirow{2}{*}{Method}} & \multicolumn{1}{c|}{\multirow{2}{*}{Shallower?}} & \multicolumn{1}{c||}{\multirow{2}{*}{\begin{tabular}[c]{@{}c@{}}Top1 \\ Acc. (\%)\end{tabular}}} & \multicolumn{2}{c|}{ 1080Ti LR (\%)} & \multicolumn{2}{c|}{ Xavier LR (\%)} \\ \cline{4-7} 
\multicolumn{1}{|l|}{} & \multicolumn{1}{c|}{} & \multicolumn{1}{c||}{} & \multicolumn{1}{c|}{bs=8} & \multicolumn{1}{c|}{bs=64} & \multicolumn{1}{c|}{bs=8} & \multicolumn{1}{c|}{bs=64} \\ \hline
\hline
\multicolumn{7}{c}{\textbf{CIFAR-10 ResNet56 baseline (93.55\%)}} \\
\hline
 Chen et al. \cite{chen2018shallowing} & \cmark & 93.09 & \textbf{26.60} & \textbf{26.31} & 26.96 & 25.66 \\
 LayerPrune$_8$-Imprint &  \cmark & \textbf{93.63} & \textbf{26.41} & \textbf{26.32} & \textbf{27.30} & \textbf{29.11} \\
 \hline
Taylor weight \cite{taylor} & \xmark & 93.15 & 0.31 & 5.28 & -0.11 & 2.67  \\
LayerPrune$_1$ & \cmark & \textbf{93.49} & 2.864 & \textbf{3.80} & 5.97 & 5.82 \\
LayerPrune$_2$ & \cmark & 93.35 & \textbf{6.46} & \textbf{8.12} & \textbf{9.33} & \textbf{11.38}  \\
 \hline
Weight norm \cite{han2015learning} & \xmark & 92.95 & -0.90 & 5.22 & 1.49 & 3.87  \\
L1 norm \cite{Li2016PruningFF} & \xmark & 93.30  & -1.09 & -0.48 &  2.31 & 1.64   \\
LayerPrune$_1$ & \cmark & \textbf{93.50} &  2.72 & 3.88 & 7.08 & 5.67  \\
LayerPrune$_2$ & \cmark & 93.39 &  \textbf{5.84} & \textbf{7.94} & \textbf{10.63} & \textbf{11.45}  \\
 \hline
Feature maps \cite{fmtaylor} & \xmark & 92.7 & -0.79 & 6.17 & 1.09 & \textbf{8.38}  \\
LayerPrune$_1$ & \cmark & \textbf{92.61} &  3.29 & 2.40 &  7.77 & 2.76 \\
LayerPrune$_2$ & \cmark & 92.28 & \textbf{6.68} & \textbf{5.63} & \textbf{11.11} & 5.05  \\
 \hline
Batch Normalization \cite{slimming} &\xmark &  93.00  & 0.6 & 3.85 & 2.26 & 1.42 \\
LayerPrune$_1$ & \cmark & \textbf{93.49} &  2.86 & 3.88 & 7.08 & 5.67 \\
LayerPrune$_2$ & \cmark & 93.35 & \textbf{6.46} & \textbf{7.94} & \textbf{10.63} & \textbf{11.31}  \\
 \hline
 LayerPrune$_{18}$-Imprint & \cmark & 92.49 & 57.31 & 55.14 & 57.57 & 63.27\\ 
 \hline
 
 
 \multicolumn{7}{c}{\textbf{CIFAR-100 ResNet56 baseline (71.2\%)}} \\
 \hline
 Chen et al. \cite{chen2018shallowing} & \cmark & 69.77 & \textbf{38.30} & 34.31 & 38.53 & 39.38  \\
LayerPrune$_{11}$-Imprint & \cmark & \textbf{71.00} & \textbf{38.68} & \textbf{35.83} & \textbf{39.52} & \textbf{54.29}  \\ 
\hline
Taylor weight \cite{taylor} &  \xmark & 71.03 & 2.13 & 5.23 & -1.1 & 3.75 \\
LayerPrune$_1$ & \cmark & \textbf{71.15} & 3.07 & 3.74 & 3.66 & 5.50 \\
LayerPrune$_2$ & \cmark & 70.82 & \textbf{6.44} & \textbf{7.18} & \textbf{7.30} & \textbf{11.00} \\
 \hline
Weight norm \cite{han2015learning} & \xmark & 71.00 & 2.52 & 6.46 & -0.3 & 3.86  \\
L1 norm \cite{Li2016PruningFF} &  \xmark & 70.65 & -1.04 & 4.06  & 0.58 & 1.34  \\
LayerPrune$_1$ & \cmark & \textbf{71.26} & 3.10 & 3.68 & 4.22 & 5.47 \\
LayerPrune$_2$ & \cmark & 71.01 & \textbf{6.59} & \textbf{7.03} & \textbf{8.00} & \textbf{10.94} \\
 \hline
Feature maps \cite{fmtaylor} & \xmark & 70.00 & 1.22 & 9.49  & -1.27 & \textbf{7.94} \\
LayerPrune$_1$ & \cmark & \textbf{71.10} & 2.81 & 3.24  & 4.46 & 5.56  \\
LayerPrune$_2$& \cmark &  70.36 & \textbf{6.06} & \textbf{6.70} & \textbf{7.72} & \textbf{7.85}  \\
 \hline
Batch Normalization \cite{slimming} & \xmark & 70.71 & 0.37 & 2.26  & -1.02 & 2.89 \\
LayerPrune$_1$ &\cmark &  \textbf{71.26} & 3.10 & 3.68 & 4.22 & 5.47 \\
LayerPrune$_2$ & \cmark & 70.97 & \textbf{6.36} & \textbf{6.78} & \textbf{7.59} & \textbf{10.94} \\ 
\hline
LayerPrune$_{18}$-Imprint & \cmark & 68.45 & 60.69 & 57.15 & 61.32 & 71.65  \\ 
 \hline
\end{tabular}}
  \caption{\textbf{Comparison of different pruning methods on ResNet56 CIFAR-10/100.} The accuracy for baseline model is shown in parentheses. LR and bs stands for latency reduction and batch size respectively. subscript $x$ in LayerPrune$_x$ indicates number of blocks removed.}
  \label{tab:cifarresnet50}
\end{table}

It is worth mentioning that the latency aware optimization ECC has an upper bound latency reduction of 11.56\%, on 1080Ti, with accuracy 16.3\%. This stems from the fact that iterative filter pruning is bounded by the network's depth and structure dependency within the network, thus not all layers are considered for pruning such as the gates at residual blocks. Besides, ECC builds a layer-wise bilinear model to approximate the latency of a model given the number of input channels and output filters per layer. This simplifies the non-linear relationship between the number of filters per layer and latency. We show the latency reduction on Xavier for an ECC pruned model optimized for 1080Ti, and this pruned model results in a latency increase on batch size 1 and the lowest latency reduction on batch size 64. This suggests that a hardware-aware filter pruned model for one hardware architecture might perform worse on another hardware than even a hardware-agnostic filter pruning method. It is worth noting that the filter pruning HRank \cite{lin2020hrank} with 2.6x FLOPs reduction shows large accuracy degradation compared to LayerPrune (71.98 vs 74.31). Even with aggressive filter pruning, speed up is noticeable with large batch size but shows small speed gain with small batch size. Within shallower models, LayerPrune outperforms SSS on the same latency budget even when SSS supports block pruning for ResNet50, which shows the effectiveness of accuracy approximation as layer importance.

\begin{table}[]
\centering
\setlength{\tabcolsep}{6pt}
\resizebox{\columnwidth}{!}{
\begin{tabular}{|@{\hspace{0.2em}}l|c|c||c|c|c|c|}
\hline
  \multicolumn{7}{|c|}{\textbf{ResNet50 baseline (76.14)}}\\
\hline
\multicolumn{1}{|@{\hspace{0.2em}}l|}{\multirow{2}{*}{Method}} & \multicolumn{1}{c|}{\multirow{2}{*}{Shallower?}} & \multicolumn{1}{c||}{\multirow{2}{*}{\begin{tabular}[c]{@{}c@{}}Top1 \\ Acc. (\%)\end{tabular}}} & \multicolumn{2}{c|}{ 1080Ti LR (\%)} & \multicolumn{2}{c|}{ Xavier LR (\%)} \\ \cline{4-7} 
\multicolumn{1}{|l|}{} & \multicolumn{1}{c|}{} & \multicolumn{1}{c||}{} & \multicolumn{1}{c|}{bs=1} & \multicolumn{1}{c|}{bs=64} & \multicolumn{1}{c|}{bs=1} & \multicolumn{1}{c|}{bs=64} \\ \hline
Weight norm \cite{han2015learning} &  \xmark & 76.50 & 6.79 & 3.46 & 6.57 & 8.06 \\
ECC \cite{ecc} &  \xmark & 75.88 & 13.52 & 1.59 & -4.91** & 3.09** \\
LayerPrune$_1$ & \cmark & \textbf{76.70} & 15.95 & 4.81 & 21.38 & 6.01 \\
LayerPrune$_2$ & \cmark & 76.52 & \textbf{20.32} & \textbf{13.23} & \textbf{26.14} & \textbf{13.20} \\
\hline

Batch Normalization &  \xmark & 75.23 & 2.49 & 1.61 & -2.79 & 4.13 \\
LayerPrune$_1$  &\cmark &  \textbf{76.70} & 15.95 & 4.81 & 21.38 & 6.01 \\
LayerPrune$_2$ & \cmark & 76.52 & \textbf{20.41} & \textbf{8.36} & \textbf{25.11} & \textbf{9.96} \\
\hline
Taylor \cite{taylor} &  \xmark & 76.4 & 2.73 & 3.6 & -1.97 & 6.60 \\
LayerPrune$_1$ & \cmark & \textbf{76.48} & 15.79 & 3.01 & 21.52 & 4.85 \\
LayerPrune$_2$ &\cmark & 75.61 & \textbf{21.35} & \textbf{6.18} & \textbf{27.33} & \textbf{8.42} \\ \hline
Feature maps \cite{fmtaylor} &  \xmark & \textbf{75.92} & 10.86 & 3.86 & 20.25 & 8.74 \\
Channel pruning* \cite{he2017channel} & \xmark &  72.26 & 3.54 & 6.13 & 2.70 & 7.42 \\
ThiNet* \cite{luo2017thinet} &  \xmark & 72.05 & 10.76 & \textbf{10.96} & 15.52 & \textbf{17.06} \\
LayerPrune$_1$ & \cmark & 75.00 & 16.56 & 2.54 & 23.82 & 4.49 \\
LayerPrune$_2$ & \cmark & 71.90 & \textbf{22.15} & 5.73 & \textbf{29.66} & 8.03 \\
 \hline
 SSS-ResNet41 \cite{huang2018data} &\cmark& 75.50 & 25.58 & 24.17 & 31.39 & 21.76 \\
 LayerPrune$_3$-Imprint &\cmark &  \textbf{76.40} & 22.63 & 25.73 & 30.44 & 20.38 \\
 LayerPrune$_4$-Imprint &\cmark &  75.82 & \textbf{30.75} & \textbf{27.64} & \textbf{33.93} & \textbf{25.43} \\
 \hline
 SSS-ResNet32 \cite{huang2018data} &\cmark &  74.20 & \textbf{41.16} & 29.69 & \textbf{42.05} & 29.59 \\
 LayerPrune$_6$-Imprint &\cmark &  \textbf{74.74} & 40.02 & \textbf{36.59} & 41.22 & 34.50 \\
 \hline
 HRank-2.6x-FLOPs* \cite{lin2020hrank} & \xmark & 71.98 & 11.89 & 36.09 & 20.63 & \textbf{40.09} \\
 LayerPrune$_7$-Imprint &\cmark & \textbf{74.31} & \textbf{44.26} & \textbf{41.01} & \textbf{41.01}  & 38.39 \\
 \hline
 
\end{tabular}}
 \caption{\textbf{Comparison of different pruning methods on ResNet50 ImageNet.} * manual pre-defined signatures. ** same pruned model optimized for 1080Ti latency consumption model in ECC optimization}
  \label{tab:imgnetresnet50}
\end{table}

\subsection{Layer pruning comparison}
In this section, we analyze different criteria for layer pruning under the same latency budget as presented in Table \ref{tab:layerpruning}. Our imprinting method consistently outperforms other methods, especially on higher latency reduction rates. Imprinting is able to get 30\% latency reduction with only 0.36\% accuracy loss from baseline. The ensemble method, although has better accuracy than the average accuracy, is still sensitive to individual errors. We further compare layer pruning by imprinting on a similar latency budget with smaller ResNet variants. We outperform ResNet34 by 1.44\% (LR=39\%) and ResNet18 by 0.56\% (LR=65\%) in accuracy showing the effectiveness of incorporating accuracy in block importance. Detailed numerical evaluation can be found in supplementary.

\begin{table}[t]
\centering
\resizebox{\columnwidth}{!}{
\begin{tabular}{|l|c|c|c|c|}
\hline
ResNet50 (76.14) & \multicolumn{4}{c|}{} \\ \hline
 & \begin{tabular}[c]{@{}c@{}}1 block \\ (LR$\approx$ 15\%)\end{tabular} & \begin{tabular}[c]{@{}c@{}}2 blocks \\ (LR$\approx$ 20\%)\end{tabular} & \begin{tabular}[c]{@{}c@{}}3 blocks \\ (LR$\approx$ 25\%)\end{tabular}  & \begin{tabular}[c]{@{}c@{}}4 blocks \\ (LR$\approx$ 30\%)\end{tabular} \\ 
 \hline
LayerPrune-Imprint & \textbf{76.72}  & \textbf{76.53} & \textbf{76.40} & \textbf{75.82} \\ \hline
LayerPrune-Taylor & 76.48 & 75.61 & 75.34 & 75.28 \\ \hline
LayerPrune-Feature map & 75.00 & 71.9 & 70.84 & 69.05  \\ \hline
LayerPrune-Weight magnitude & \textbf{76.70} & \textbf{76.52} & 76.12 & 74.33 \\ \hline
LayerPrune-Batch Normalization & \textbf{76.70} & 76.22 & 75.84 & 75.03  \\ \hline
LayerPrune-Ensemble & \textbf{76.70} & 76.11 & 75.76 & 75.01 \\ \hline
\end{tabular}}

\caption{\textbf{Comparison of different layer pruning methods supported by LayerPrune on ResNet50 ImageNet.} Latency reduction is calculated on 1080Ti with batch size 1.}
\label{tab:layerpruning}
\end{table}

\section{Conclusion} We presented LayerPrune framework which includes a set of layer pruning methods. We show the benefits of LayerPrune on latency reduction compared to filter pruning. The key findings of this paper are the following:
\begin{itemize}
  \item For a filter criterion, training a LayerPrune model based on this criterion achieves the same, if not better, accuracy as the filter pruned model obtained by using the same criterion.
  \item Filter pruning compresses the number of convolution operations per layer and thus latency reduction depends on hardware architecture, while LayerPrune removes the whole layer. As result, filter pruned models might produce non-optimal matrix shapes for the compute kernels that can lead even to latency increase on some hardware targets and batch sizes. 
  \item Filter pruned models within a latency budget have a larger variance in accuracy than LayerPrune. This stems from the fact that the relation between latency and number of filters is non-linear and optimization constrained by a resource budget requires complex per-layer pruning ratios selection.
  \item We also showed the importance of incorporating accuracy approximation in layer ranking by imprinting.
\end{itemize}
 
\section*{Acknowledgment} We thank Compute Canada and WestGrid for their supercomputers to conduct our experiments.
 

  \label{fig:oracle}


\bibliographystyle{splncs}
\bibliography{egbib}

\beginsupplement

\section*{Appendix}
\section{CIFAR}
\subsection{Training setup}
We follow standard hyperparameters used for fine-tuning \cite{taylor,liu2018rethinking,huang2018data}: 30 epoch learning rate 1e$^{-3}$ on SGD optimizer. That is except comparison with Chen et al. \cite{chen2018shallowing} as the authors train the pruned model with the standard hyperparameters used for training from scratch: 160 epoch with initial learning rate 0.1 and decays on epoch [81, 122] by 0.1. \\
\textbf{Layer pruning:} For layer pruning, we calculate layer importance as explained in the paper using a one-shot pass over the training set. \\
\textbf{Filter pruning:} For filter pruning, we prune total 500 and 100 filters in VGG19 and ResNet56 respectively for global-based filter importance criteria such as weight norm, Taylor, Feature maps. We follow the same iterative pruning hyperparameter setup as Taylor \cite{taylor}. We prune 100 filters each 10 minibatches. Other pruning methods, we report results using their published code with default setup setting such as slimming \cite{slimming} and ECC\cite{ecc}. \\

\subsection{Ablation}

\subsubsection{Number of filters pruned.} 
\begin{figure}
\begin{subfigure}{.5\textwidth}
  \centering
  \includegraphics[width=\linewidth]{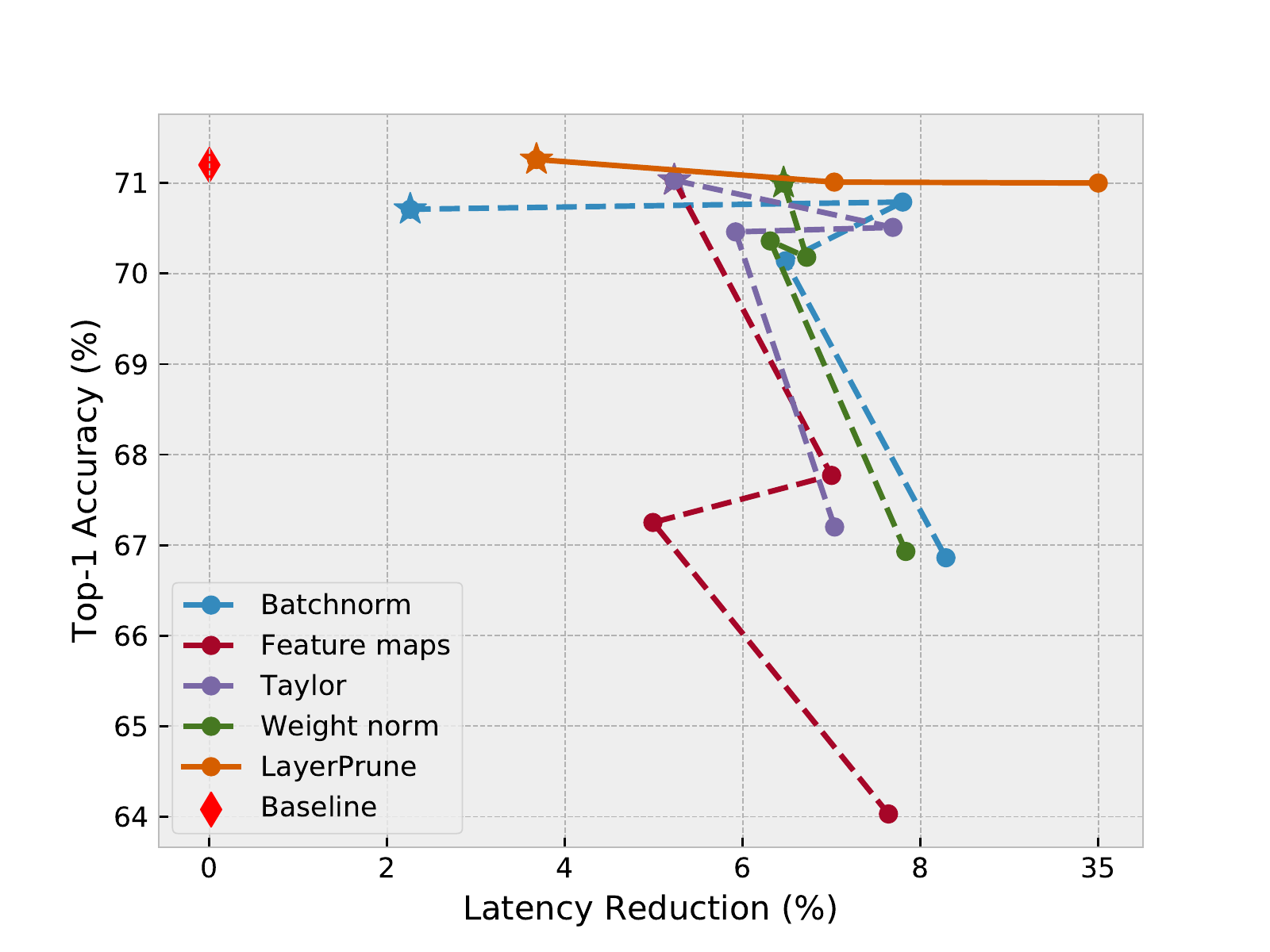}
  \caption{CIFAR-100/ResNet56}
  \label{fig:c10wn}
\end{subfigure}
\begin{subfigure}{.5\textwidth}
  \centering
  \includegraphics[width=\linewidth]{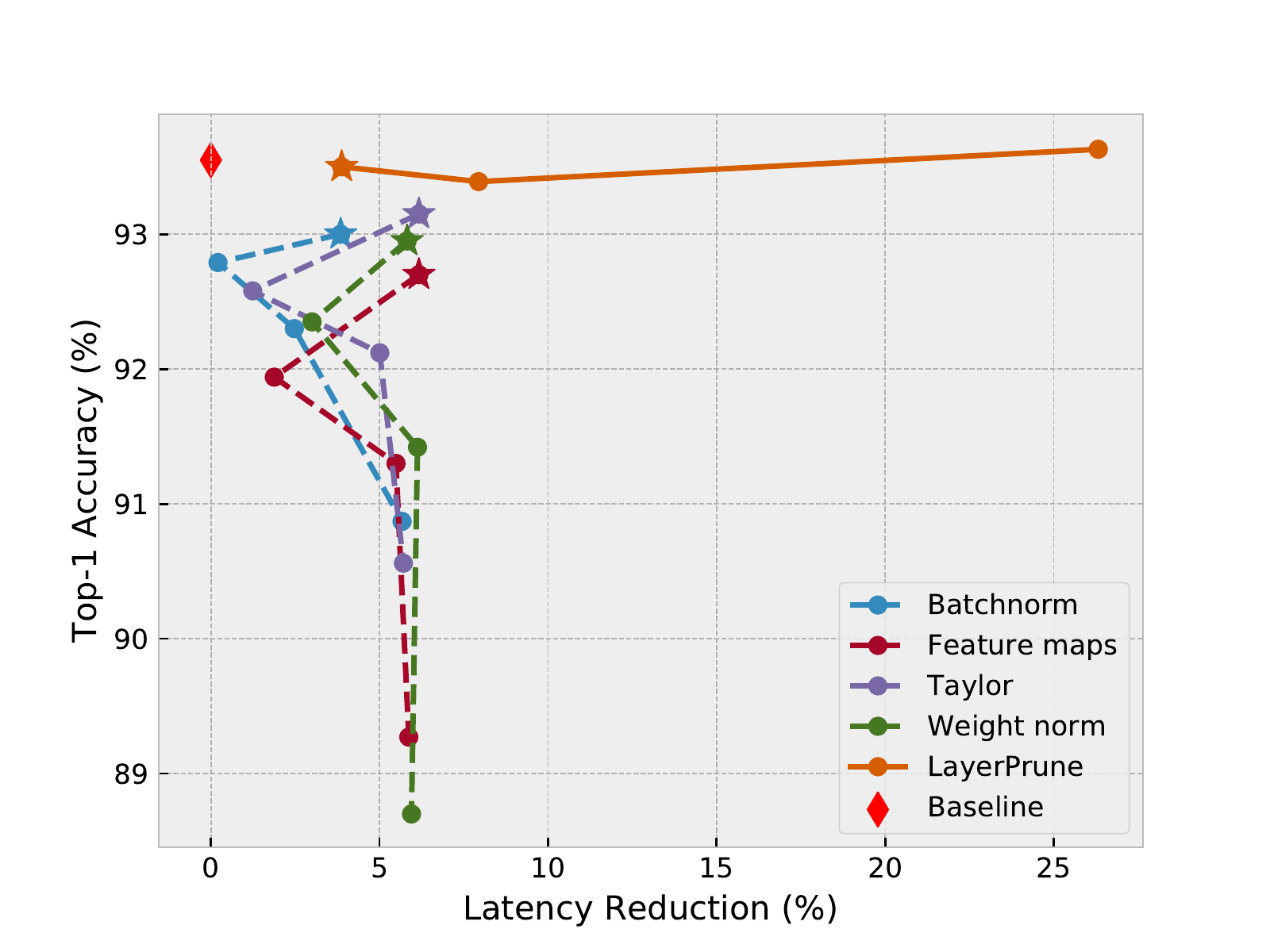}
  \caption{CIFAR-10/ResNet56}
  \label{fig:c10wn}
\end{subfigure}
\begin{subfigure}{\textwidth}
  \centering
  \includegraphics[width=0.5\linewidth]{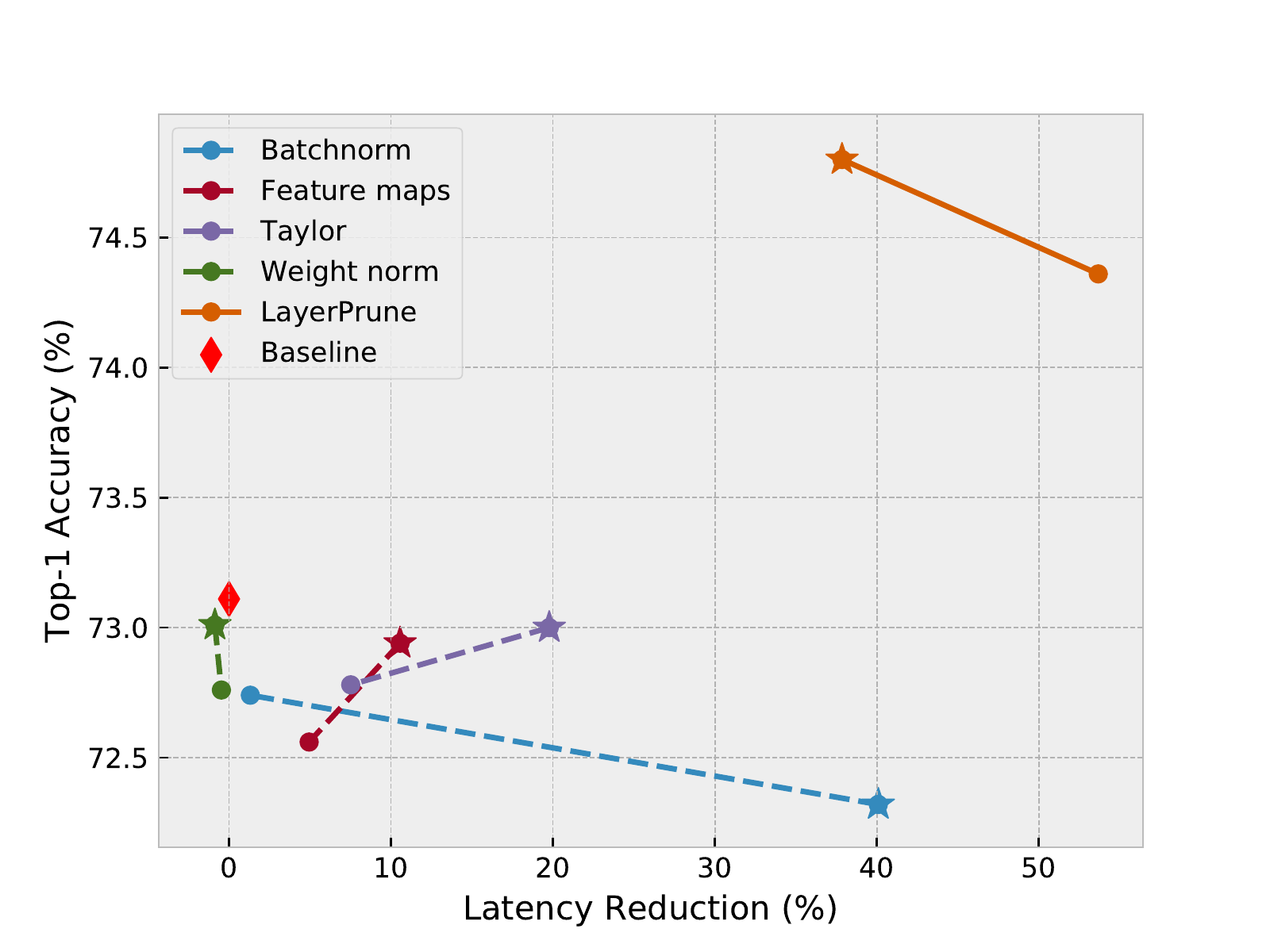}
  \caption{CIFAR-100/VGG19-BN}
  \label{fig:c10wn}
\end{subfigure}
\caption{Latency reduction of different filter pruning methods under different pruning ratios. Star in each method indicates the lowest pruning ratio (starting point). Dots are connected based on ascending order of number of filters pruned.}
\label{fig:hyper}
\end{figure}

We show accuracy degradation on aggressive filter pruning and the achieved latency reduction compared to LayerPrune. \textbf{Fig. \ref{fig:hyper}} shows filter pruning under different number of filters pruned (i.e 100:400) and latency reduction on GPU 1080Ti on batch size=64. Dots are connected based on ascending order of number of filters pruned. It is apparent that pruning more filters doesn't necessarily decrease latency and the relationship between pruned filters and latency reduction is non-linear. In CIFAR-100, latency reduction $\approx$ 8\% results in large drop in accuracy from 71.2\% to 67\%. It is worth noting that LayerPrune is able to achieve up to 35\% latency reduction with accuracy 71\%. Similarly on CIFAR-10, pruning 50\% of the filters can only achieve around 5\% latency reduction. \\
As for VGG19, the maximally achieved pruning latency reduction is 20\% to maintain the accuracy from baseline; while LayerPrune finds better models than baseline and filter pruned methods. On comparison with the random experiment shown in Section (4.2) in main paper, filter pruning methods hover around baseline accuracy and fails to discover other regularized models compared to layer pruning.

\subsubsection{One-shot vs iterative.} We also conducted experiments on one-shot vs iterative filter pruning to be comparable with our one-shot LayerPrune pruning step. Our reported results on iterative filter pruning follows the same setup used in literature \cite{taylor}, that is prune 10\% each pruning iteration. \textbf{Table \ref{tab:iterative}} shows results of iterative vs one-shot (pruning total number of filters at once). Consistent with \cite{taylor,liu2018rethinking}, iterative pruning (i.e re-evaluating criterion of filter after each prune) gives a slightly better accuracy. That shows that it is mandatory for filter pruning to be iterative.

\begin{table}[]
\centering
\begin{tabular}{|l|l|l|l|l|}
\hline
Dataset/Model & Pruning ratio & Criterion & Iterative & One-shot \\ \hline
\multirow{10}{*}{CIFAR100/ResNet56 (71.20\%)} & \multirow{5}{*}{20\%} & Weight Norm & 70.18 & 70.00 \\ \cline{3-5} 
 &  & Feature Maps & 67.77 & 67.7 \\ \cline{3-5} 
 &  & Taylor & 70.51 & 70.01 \\ \cline{3-5} 
 &  & Batchnorm & 70.79 & 70.36 \\ \cline{3-5} 
 &  & \textbf{Median} & \textbf{70.34} & 70.00 \\ \cline{2-5} 
 & \multirow{5}{*}{30\%} & Weight Norm & 70.36 & 69.1 \\ \cline{3-5} 
 &  & Feature Maps & 67.25 & 66.34 \\ \cline{3-5} 
 &  & Taylor & 70.46 & 68.27 \\ \cline{3-5} 
 &  & Batchnorm & 70.14 & 69.4 \\ \cline{3-5} 
 &  & \textbf{Median} & \textbf{70.25} & 68.68 \\ \hline
\multirow{5}{*}{CIFAR10/ResNet56 (93.55\%)} & \multirow{5}{*}{20\%} & Weight Norm & 92.35 & 92.31 \\ \cline{3-5} 
 &  & Feature Maps & 91.94 & 91.9 \\ \cline{3-5} 
 &  & Taylor & 92.88 & 92.8 \\ \cline{3-5} 
 &  & Batchnorm & 92.79 & 92.76 \\ \cline{3-5} 
 &  & \textbf{Median} & \textbf{92.57} & 92.53 \\ \hline
\end{tabular}
\caption{Evaluation of iterative and one-shot filter pruning. Baseline accuracy indicates in parentheses.}
\label{tab:iterative}
\end{table}

To analyze the sensitivity of ranking by imprinting on layer pruned models, we calculated Spearman's rank-order correlation between layer ranking by one-shot and layer ranking by re-calculating ranks iteratively after each pruning step. \textbf{Table \ref{tab:literative}} shows accuracy of one-shot and iterative layer pruning and their ranking correlation. The Spearman column indicates high positive relationship between both ranking methods demonstrating the robustness of ranking by imprinting. We observed the difference in ranking is between similarly important layers and this explains why accuracy isn't significantly affected even as correlation decreases, and it shows the sufficiency of one-shot rank estimation with imprinting.\newline

\begin{table}[]
\centering
\begin{tabular}{|l|c|c|c|c|c|c|}
\hline
 & \multicolumn{3}{c|}{CIFAR-10 ResNet-56 (93.55\%)} & \multicolumn{3}{c|}{CIFAR-100 ResNet-56 (71.2\%)} \\ \hline
N pruned & \multicolumn{1}{l|}{One-shot (\%)} & \multicolumn{1}{l|}{Iterative (\%)} & \multicolumn{1}{l|}{Spearman} & \multicolumn{1}{l|}{One-shot (\%)} & \multicolumn{1}{l|}{Iterative (\%)} & \multicolumn{1}{l|}{Spearman} \\ \hline
1 & 93.32 & 93.32 & 0.99 & 71.10 & 71.10 & 0.96 \\ \hline
2 & 93.28 & 93.31 & 0.97 & 70.93 & 70.94 & 0.97 \\ \hline
3 & 93.17 & 93.15 & 0.96 & 70.88 & 70.86 & 0.94 \\ \hline
4 & 93.10 & 93.03 & 0.96 & 70.64 & 70.71 & 0.91 \\ \hline
\end{tabular}
\caption{Spearman rank correlation between one-shot and iterative ranking with imprinting.}
\label{tab:literative}
\end{table}

\subsection{Architectures}
Architectures for results on VGG19-BN are presented in \textbf{Table \ref{tab:cifar100vggarch}}. All layer pruning methods mostly agree on removing same layers. While in filter pruning methods, as minimum number of filters are required per layer, the early layers are pruned as well and thus hurting accuracy.

\begin{table}[]
\centering
\resizebox{\columnwidth}{!}{
\begin{tabular}{l|l|l}
\hline
Method          & Accuracy &  Architecture      \\
\hline
VGG19 (baseline)   &    73.11      &{[}64, 64, 'M', 128, 128, 'M', 256, 256, 256, 256, 'M', 512, 512, 512, 512, 'M', 512, 512, 512, 512, 'M'{]} \\
\hline
Weight norm \cite{han2015learning}   & 73.01 & {[}47, 64, 'M', 127, 128, 'M', 256, 256, 256, 256, 'M', 512, 508, 494, 472, 'M', 502, 512, 499, 509, 'M'{]} \\
ECC \cite{ecc} & 72.71    & {[}50, 23, 'M', 128, 128, 'M', 254, 254, 254, 254, 'M', 508, 311, 164, 131, 'M', 158, 319, 509, 64, 'M'{]} \\
Layer pruning$_2$   & 73.60     & {[}64, 64, 'M', 128, 128, 'M', 256, 256, 256, 256, 'M', 512, 512, 512, 512, 'M', \textbf{0, 0,} 512, 512,'M'{]}      \\
Layer pruning$_5$ & 74.80     & {[}64, 64, 'M', 128, 128, 'M', 256, 256, 256, 256, 'M', 512, 512, \textbf{0, 0,} 'M', \textbf{0, 0, 0,} 512, 'M'{]}               \\
\hline
Slimming \cite{slimming}   & 72.32 & {[}42, 64, 'M', 125, 128, 'M', 255, 256, 255, 256, 'M', 433, 291, 82, 46, 'M', 45, 44, 62, 367, 'M'{]}      \\
Layer pruning$_2$ & 73.60     & {[}64, 64, 'M', 128, 128, 'M', 256, 256, 256, 256, 'M', 512, 512, 512, 512, 'M', \textbf{0, 0,} 512, 512,'M'{]}       \\
Layer pruning$_5$ & 74.80     & {[}64, 64, 'M', 128, 128, 'M', 256, 256, 256, 256, 'M', 512, 512, \textbf{0, 0,} 'M', \textbf{0, 0, 0,} 512, 'M'{]}               \\
\hline
Taylor \cite{taylor} & 72.61    & {[}61, 64, 'M', 127, 128,'M', 256, 256, 256, 256,'M', 512, 505, 383, 205,'M', 109, 118, 422,482, 'M'{]}    \\
Layer pruning$_2$   & 73.60     & {[}64, 64, 'M', 128, 128, 'M', 256, 256, 256, 256, 'M', 512, 512, 512, \textbf{0,} 'M', \textbf{0,} 512, 512, 512,'M'{]}          \\
Layer pruning$_5$  & 74.80     & {[}64, 64, 'M', 128, 128, 'M', 256, 256, 256, 256, 'M', 512, 512, \textbf{0, 0}, 'M', \textbf{0, 0, 0,} 512, 'M'{]}    \\
\hline
\end{tabular}}
 \caption{Architectures of different pruning methods on VGG19-BN CIFAR-100. $x$ in Layer pruning$_x$ indicates number of layers removed. Number of filters per layer is shown where 0 indicates removed layers and 'M' indicates max pooling operation.}
  \label{tab:cifar100vggarch}
\end{table}

\begin{figure}
\begin{subfigure}{.5\textwidth}
  \centering
  \includegraphics[width=\linewidth]{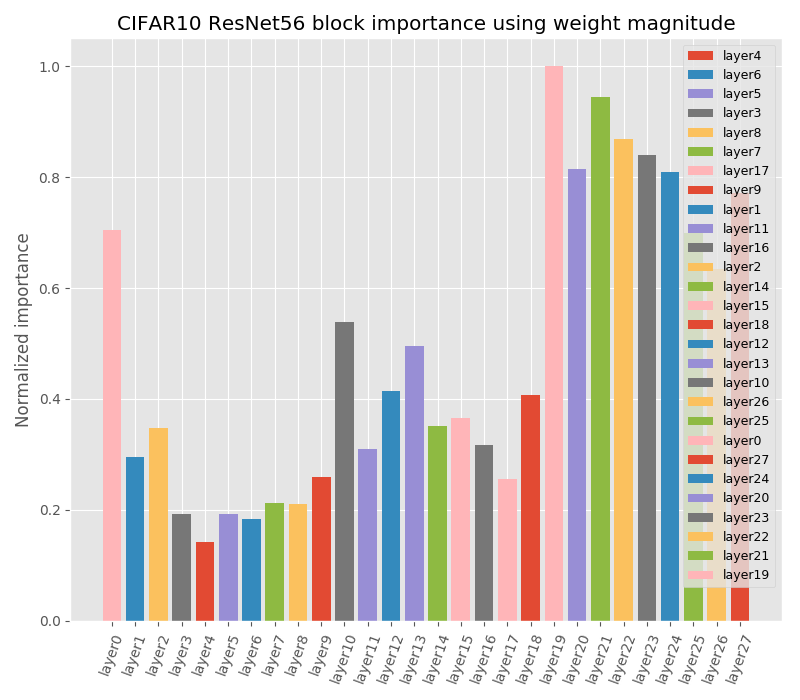}
  \caption{Weight norm}
  \label{fig:c10wn}
\end{subfigure}
\begin{subfigure}{.5\textwidth}
  \centering
  \includegraphics[width=\linewidth]{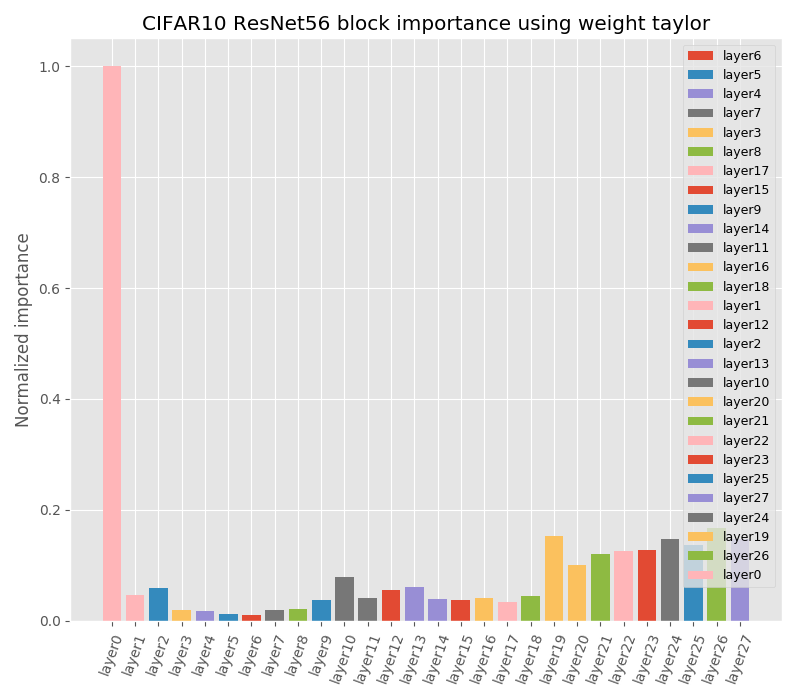}
  \caption{Weight Taylor}
  \label{fig:c10wn}
\end{subfigure}
\begin{subfigure}{.5\textwidth}
  \centering
  \includegraphics[width=\linewidth]{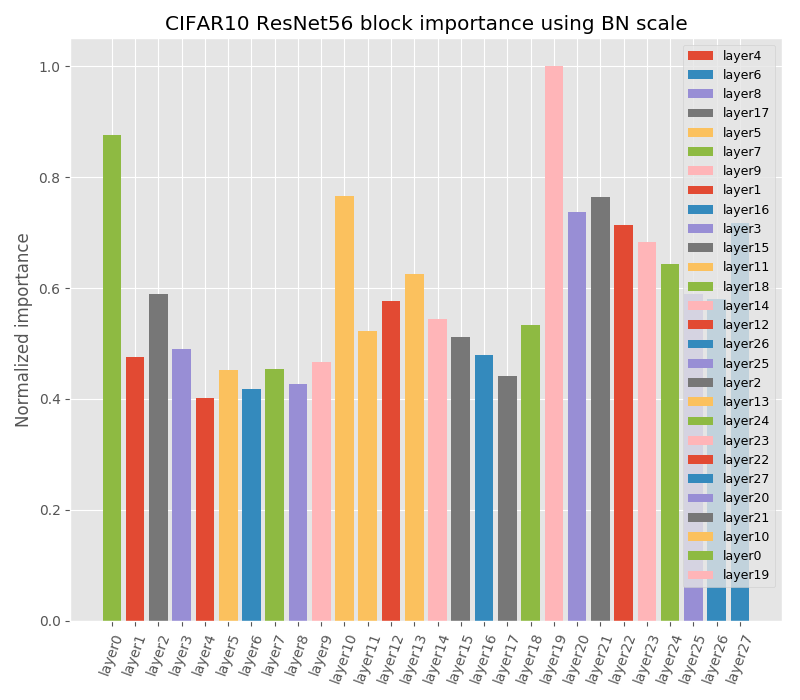}
  \caption{Batch Normalization}
  \label{fig:c10bn}
\end{subfigure}
\begin{subfigure}{.5\textwidth}
  \centering
  \includegraphics[width=\linewidth]{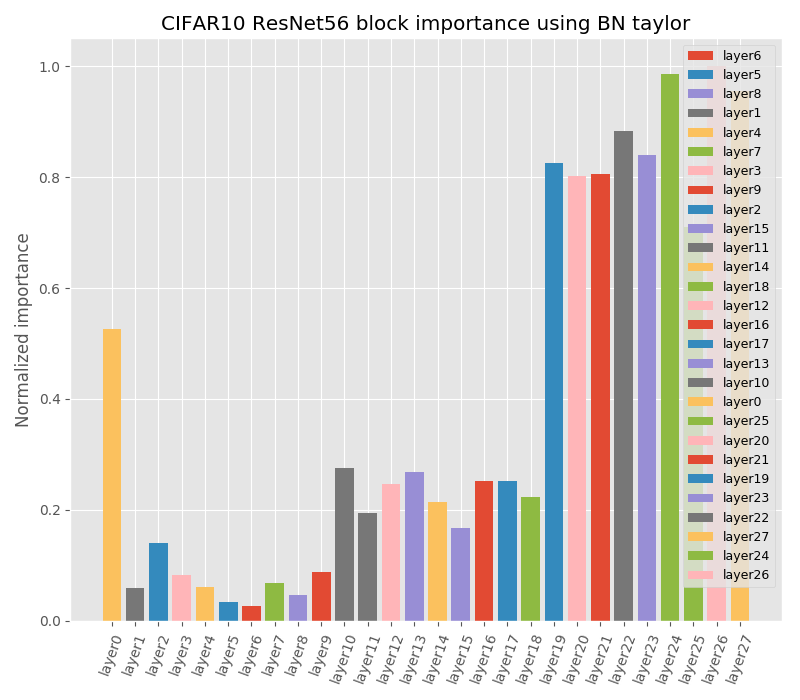}
  \caption{Batch Normalization Taylor}
  \label{fig:c10bnt}
\end{subfigure}
\begin{subfigure}{.5\textwidth}
  \centering
  \includegraphics[width=\linewidth]{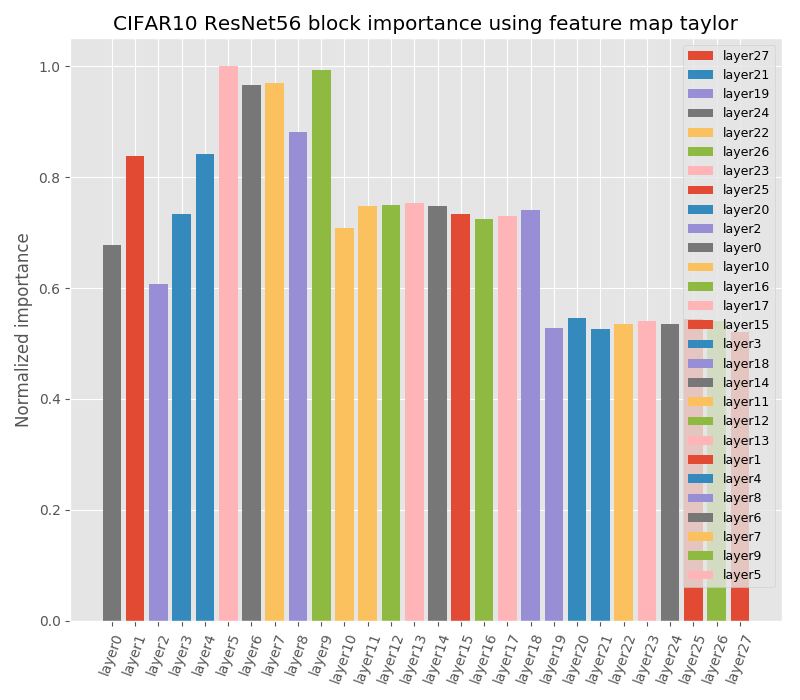}
  \caption{Feature maps}
  \label{fig:c10fm}
\end{subfigure}
\begin{subfigure}{.5\textwidth}
  \centering
  \includegraphics[width=\linewidth]{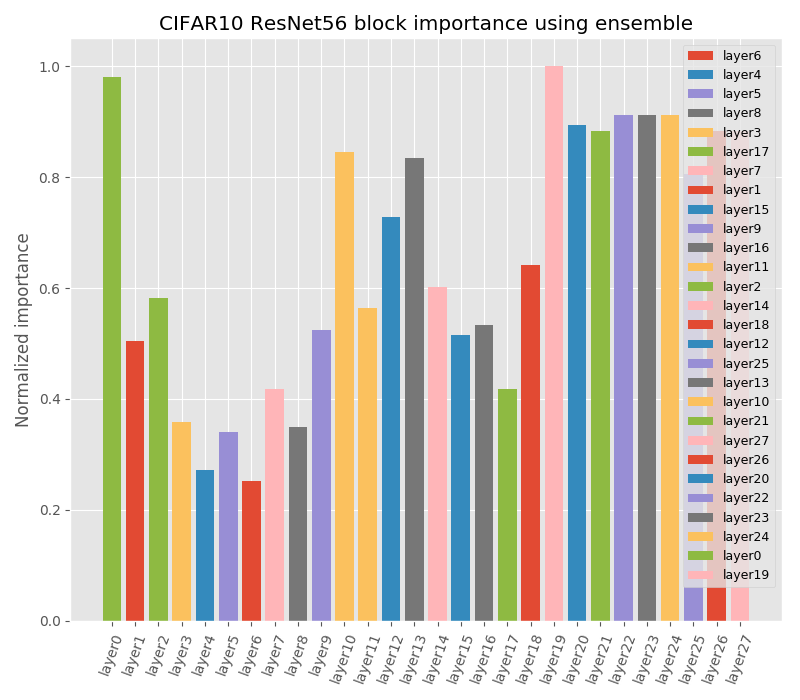}
  \caption{Ensemble}
  \label{fig:c10ensemble}
\end{subfigure}
\caption{Plots of block importance using different layer criterion on CIFAR-10 ResNet56. Legend on each sub-plot shows sorted blocks in ascending order based on importance.}
\label{fig:c10bars}
\end{figure}

\begin{figure}
\begin{subfigure}{.5\textwidth}
  \centering
  \includegraphics[width=\linewidth]{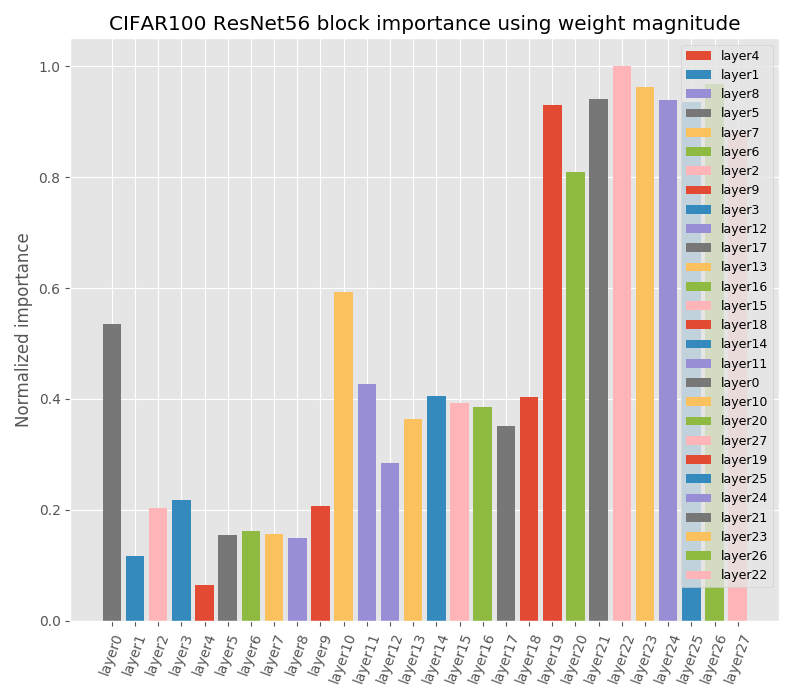}
  \caption{Weight norm}
  \label{fig:c100wn}
\end{subfigure}
\begin{subfigure}{.5\textwidth}
  \centering
  \includegraphics[width=\linewidth]{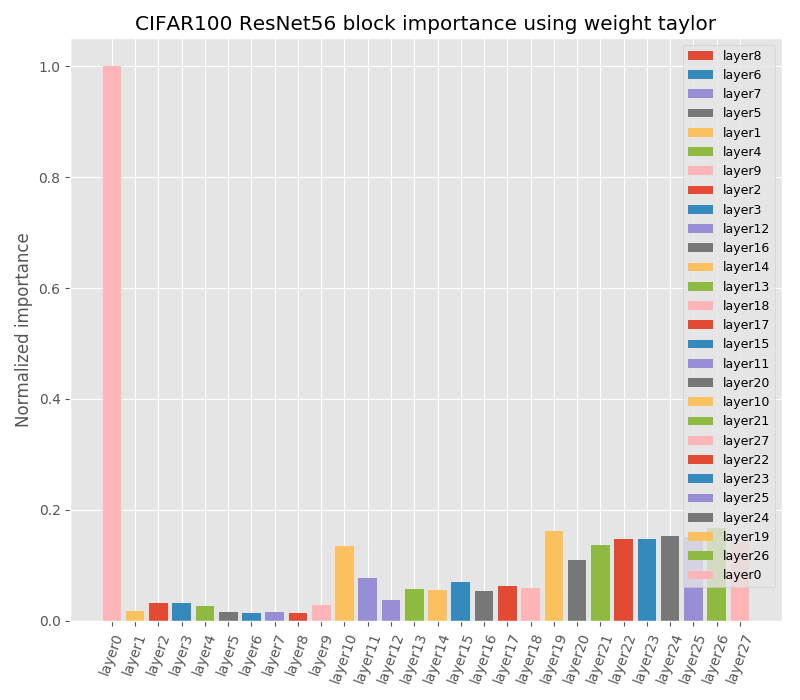}
  \caption{Weight Taylor}
  \label{fig:c100wn}
\end{subfigure}
\begin{subfigure}{.5\textwidth}
  \centering
  \includegraphics[width=\linewidth]{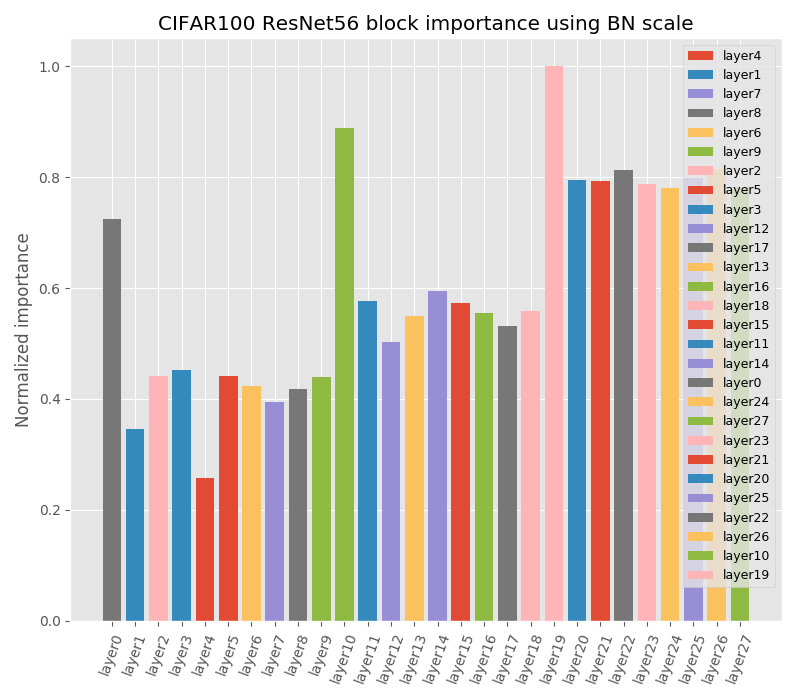}
  \caption{Batch Normalization}
  \label{fig:c100bn}
\end{subfigure}
\begin{subfigure}{.5\textwidth}
  \centering
  \includegraphics[width=\linewidth]{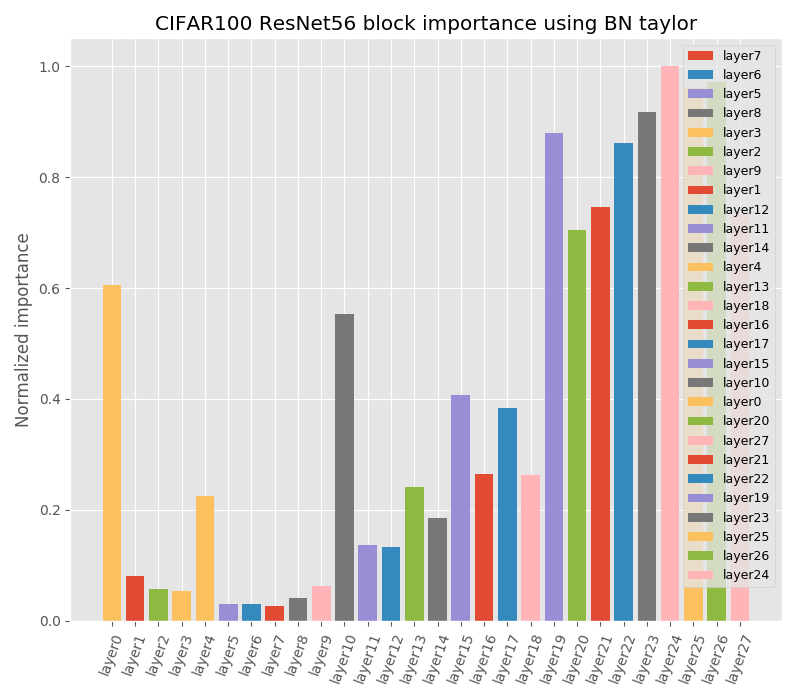}
  \caption{Batch Normalization Taylor}
  \label{fig:c100bnt}
\end{subfigure}
\begin{subfigure}{.5\textwidth}
  \centering
  \includegraphics[width=\linewidth]{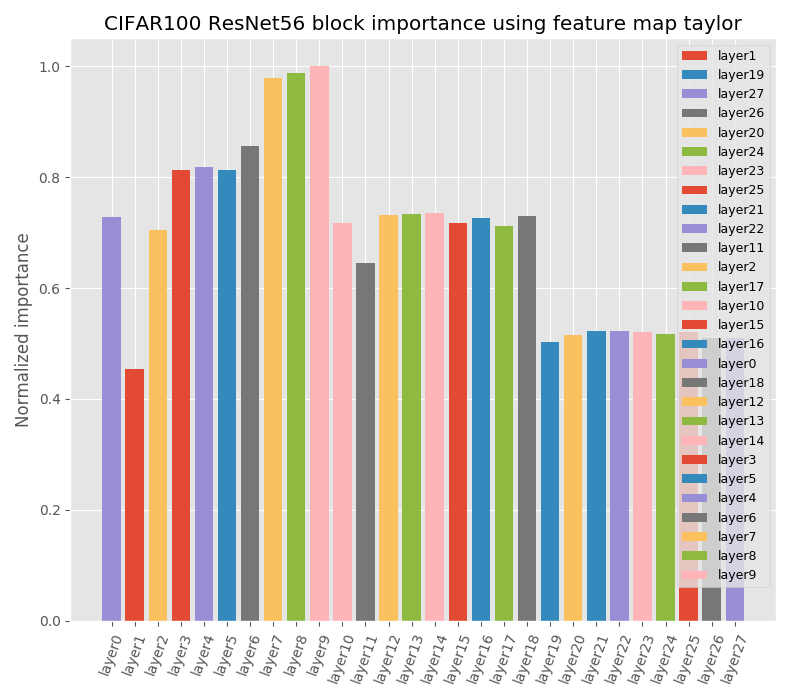}
  \caption{Feature maps}
  \label{fig:c100fm}
\end{subfigure}
\begin{subfigure}{.5\textwidth}
  \centering
  \includegraphics[width=\linewidth]{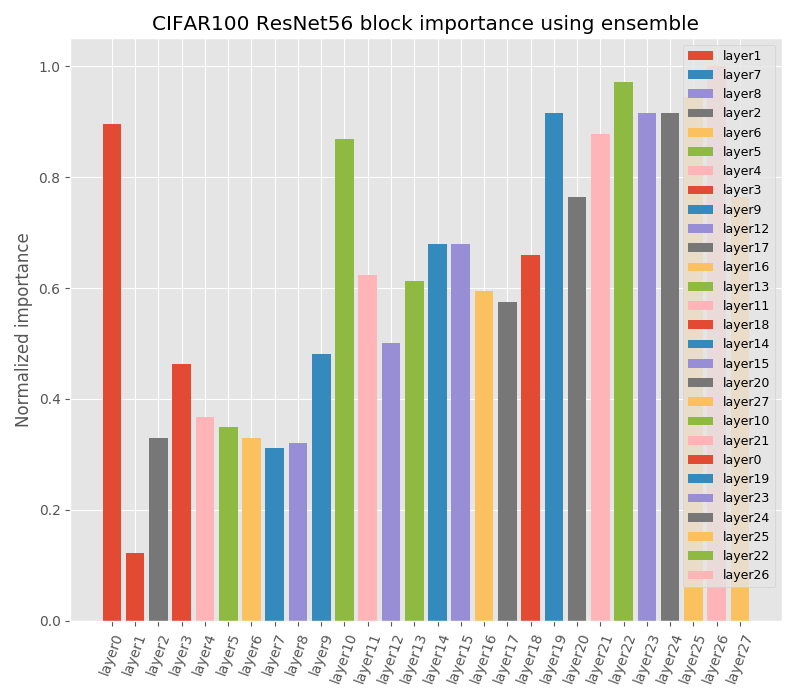}
  \caption{Ensemble}
  \label{fig:c100ensemble}
\end{subfigure}
\caption{Plots of block importance using different layer criterion on CIFAR-100 ResNet56. Legend on each sub-plot shows sorted blocks in ascending order based on importance.}
\label{fig:c100bars}
\end{figure}

\textbf{ResNet56} has 3 groups of 9 basicblocks where each basicblock has two 3x3 convolution layer. We show block importance based on each criterion for CIFAR-10 in Fig. \ref{fig:c10bars} and CIFAR-100 in \textbf{Fig. \ref{fig:c100bars}}. Weight magnitude, Batch Normalization and Taylor magnitude criteria have similar block ordering that focus more on the early layers. On the other hand, feature maps criterion is more biased to pruning the deeper layers. This stems from the fact that as we go deeper, feature maps tend to be sparser and so their importance calculated using Taylor on feature maps \cite{fmtaylor} will lead to a bias and failure in deeper models. Ensemble selects layers that are constantly voted as not important (e.g CIFAR-10 blocks 6,4,5), however, it is sensitive to individual errors. For example in CIFAR-10, ensemble prioritizes pruning block17 over block7 even when the latter has lower ranks in most of the criteria but the large ranking gap in one criterion, that is feature maps criterion, resulted in block17 to have lower rank.

\clearpage

\section{ImageNet}
\subsection{Training setup}
\textbf{Filter pruning.} We follow the same setup as Taylor \cite{taylor} for global-based filter pruning, we prune 100 filters each 30 minibatches for 10 iterations. For methods like ECC \cite{ecc}, slimming \cite{slimming}, SSS \cite{huang2018data} and HRank \cite{lin2020hrank}, we report using their default hyperparameter either reported in their papers or using their code. We fine-tune using the same setup as previously mentioned in CIFAR. 

\subsection{Ablation}

\subsubsection{Aggressive layer pruning.} In this section we show results on small light-weight models with accuracy drop from baseline to show the effectiveness of LayerPrune under high pruning ratio. We compare pruned models from ResNet50 and ResNet34 with handcrafted small variants of ResNet. Results are presneted in \textbf{Table \ref{tab:aggressive}}. We achieve 1.44\% better accuracy on similar latency as ResNet34 (74.74 vs 73.30). In addition, we match ResNet34 accuracy (73.39 vs 73.30) with 1.2x speedup. Similarly, we match ResNet18 latency with 0.5\% higher accuracy.

\begin{table}[]
\centering
\resizebox{0.7\columnwidth}{!}{
\begin{tabular}{|l|c|c|c|}
\hline
Model & Accuracy & FPS (1080Ti) & FPS (Xavier) \\
\hline
ResNet50 & 76.14 & 129 & 62 \\
\hline
LayerPrune$_6$-ResNet50 & \textbf{74.74} & 214 & 108\\ 
LayerPrune$_7$-ResNet50 & 74.31 & 239 & 114 \\ 
LayerPrune$_8$-ResNet50 & 73.39 & \textbf{248} & \textbf{122}\\ 
ResNet34 & 73.30 & 206 & 105 \\ 
\hline
LayerPrune$_8$-ResNet34 & \textbf{70.32} & 364 & 168 \\
LayerPrune$_9$-ResNet34 & 69.00 & \textbf{405} & \textbf{181} \\
ResNet18 & 69.76 & 360 & 169 \\
\hline
\end{tabular}}
 \caption{Accuracy with small handcrafted ResNets on similar frames per second.}
  \label{tab:aggressive}
\end{table}

\subsubsection{Training from scratch} Traning from scratch for ImageNet is done for 90 epochs with $0.1$ initial lr, $0.1$ lr decay each 30 epochs. Fine-tuning is done for 30 epochs with 1e$-3$ initial lr, $0.1$ lr decay each 10 epoch. In \textbf{Table \ref{tab:scratch}}, we compare our LayerPrune models trained from scratch and fine-tuned. Fine-tuned models consistently outperform training from scratch of the same pruned architecture.

\begin{table}
\centering
\resizebox{0.45\columnwidth}{!}{
\begin{tabular}{|l|l|l|}
\hline
N Blocks pruned & Fine-tuned & Scratch \\
\hline
1 & \textbf{76.72} & 75.70 \\ \hline
2 & \textbf{76.53} & 75.96 \\ \hline
3 & \textbf{76.40} & 75.80 \\ \hline
4 & \textbf{75.82} & 75.0 \\ \hline
\end{tabular}}
 \caption{Accuracy of our ResNet50 pruned models trained from scratch and fine-tuned.}
  \label{tab:scratch}
\end{table}

\subsubsection{Training speed} End-to-end optimization filter pruning methods such as slimming require training from scratch with sparsity inducing terms in the training. This requires 90 epochs in ImageNet. All methods, including ours, fine-tuned for 30 epochs. Hence, our layer-pruning is 4 times faster than these methods.\\
For iterative filter pruning methods we observed an average 1.9x speedup in the fine-tuning phase in layer-pruned models compared to fine-tuning phase in filter pruning. The training is conducted on 4 x V100 GPUs.

\subsection{Architectures}
\begin{figure}
\begin{subfigure}{.5\textwidth}
  \centering
  \includegraphics[width=\linewidth]{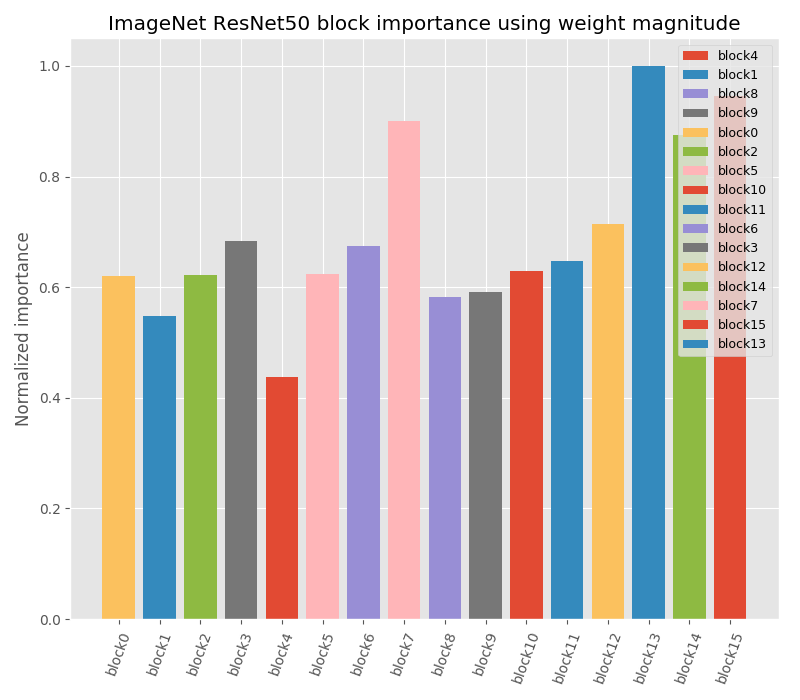}
  \caption{Weight norm}
  \label{fig:c100wn}
\end{subfigure}
\begin{subfigure}{.5\textwidth}
  \centering
  \includegraphics[width=\linewidth]{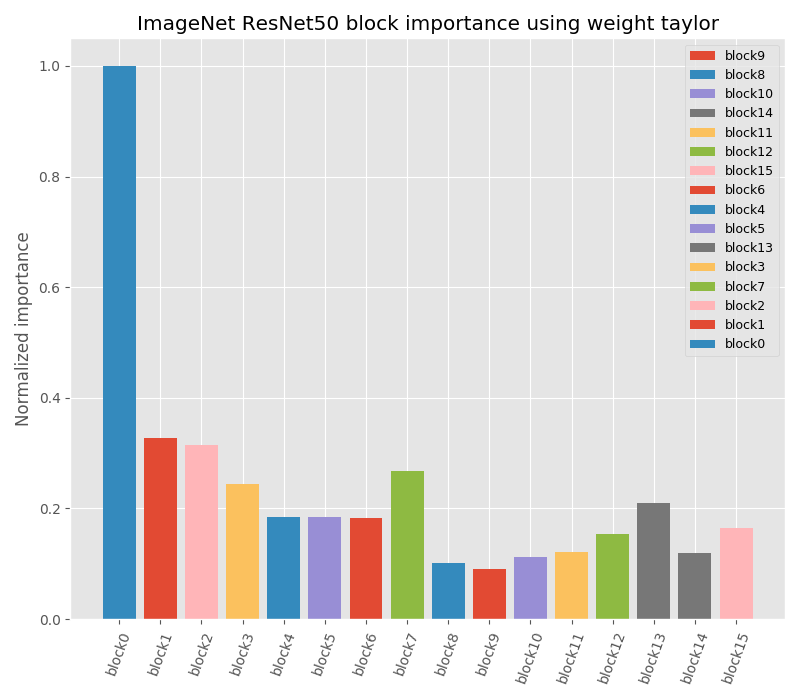}
  \caption{Weight Taylor}
  \label{fig:c100wn}
\end{subfigure}
\begin{subfigure}{.5\textwidth}
  \centering
  \includegraphics[width=\linewidth]{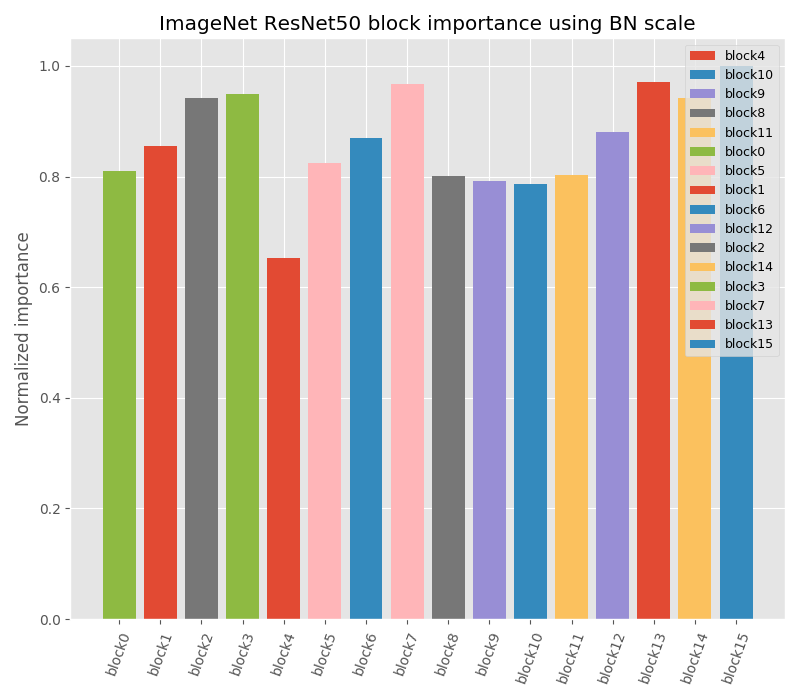}
  \caption{Batch Normalization}
  \label{fig:c100bn}
\end{subfigure}
\begin{subfigure}{.5\textwidth}
  \centering
  \includegraphics[width=\linewidth]{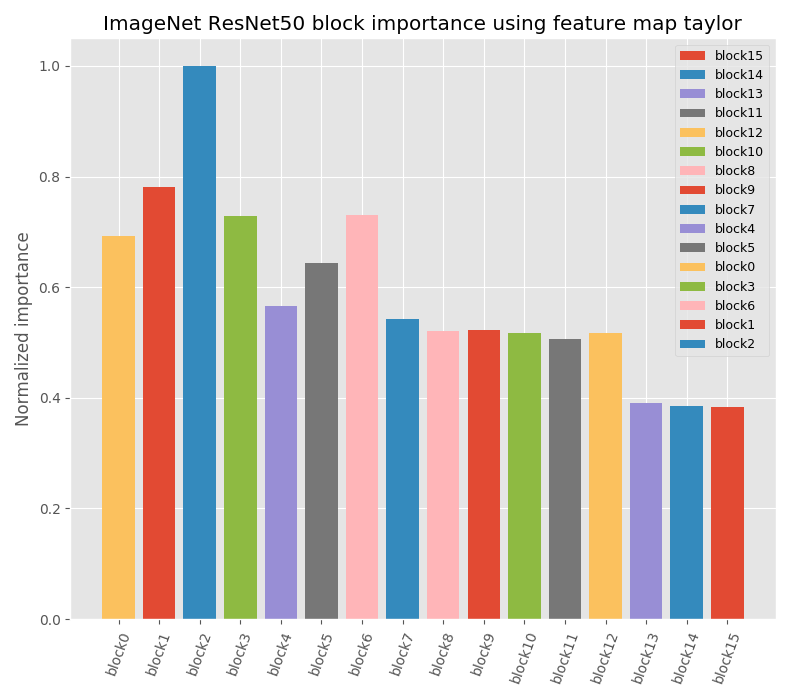}
  \caption{Feature maps}
  \label{fig:c100fm}
\end{subfigure}
\begin{subfigure}{.5\textwidth}
  \centering
  \includegraphics[width=\linewidth]{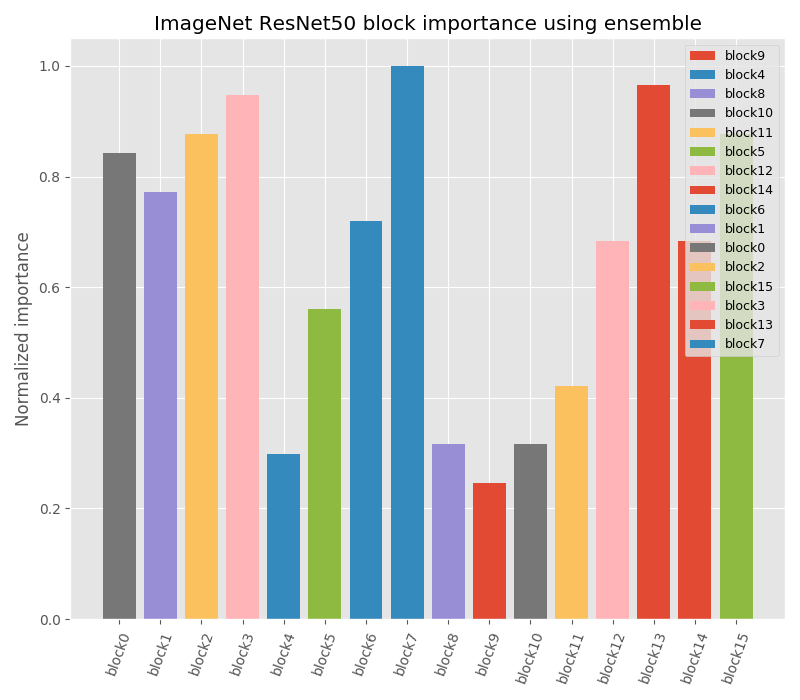}
  \caption{Ensemble}
  \label{fig:c100ensemble}
\end{subfigure}
\begin{subfigure}{.5\textwidth}
  \centering
  \includegraphics[width=\linewidth]{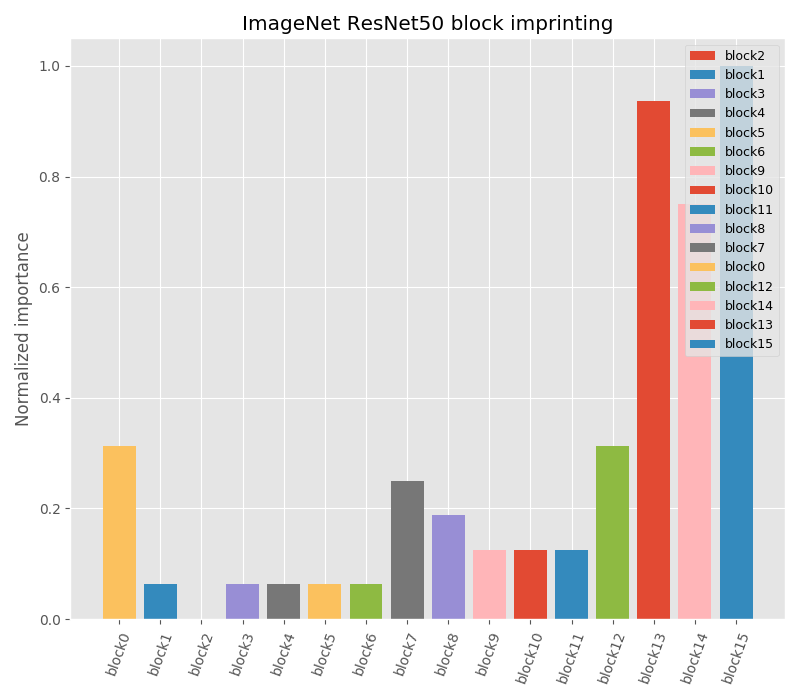}
  \caption{Imprinting}
  \label{fig:c100ensemble}
\end{subfigure}
\caption{Plots of block importance (normalized for visualization) using different layer criterion on ImageNet ResNet50. Legend on each sub-plot shows sorted blocks in ascending order based on importance.}
\label{fig:imgnetresnet}
\end{figure}

\textbf{ResNet50} Details of block importance for each criterion is shown in \textbf{Fig. \ref{fig:imgnetresnet}}. Unlike criteria that depend on filter statistics like weight values or gradients, imprinting asses quality of a block based on accuracy gained. This handles the network's distribution discrepancy in these statistics.

%
%
\clearpage

\end{document}